%% file: 1_paper.tex
\documentclass{article}

\usepackage[final]{neurips_2021}

\usepackage[utf8]{inputenc} %
\usepackage[T1]{fontenc}    %
\usepackage{hyperref}       %
\usepackage{xcolor}
\hypersetup{
  colorlinks,
  linkcolor={red!50!black},
  citecolor={blue!50!black},
  urlcolor={blue!80!black}
  }
  \definecolor{orange}{HTML}{ff7f0e}
  \definecolor{blue}{HTML}{1f77b4}
  
\usepackage{url}            %
\usepackage{booktabs}       %
\usepackage{amsfonts}       %
\usepackage{nicefrac}       %
\usepackage{microtype}      %
\usepackage{comment}     
\usepackage{amssymb}
\usepackage{amsmath,amsfonts,bm}
\usepackage{nicefrac}
\usepackage{textcomp}
\usepackage[capitalise]{cleveref}
\usepackage{enumitem}%
\usepackage{subcaption}
\usepackage{floatrow}
\newfloatcommand{capbtabbox}{table}[][0.48\textwidth]
\usepackage{tabularx}
\usepackage{makecell}

\title{Variational Diffusion Models}

\author{%
  Diederik P. Kingma$^*$, Tim Salimans$^*$, Ben Poole, Jonathan Ho\\
  Google Research\\
}

\input{2_defs_general.tex}

\input{3_defs_custom.tex}
\newcommand{\bsst}{\bsigma_{t|s}}
\newcommand{\lT}{\mathcal{L}_T(\rvx)}
\newcommand{\ldT}{\mathcal{L}_{2T}(\rvx)}
\newcommand{\linfty}{\mathcal{L}_{\infty}(\rvx)}

\newcommand{\snr}{\text{SNR}}
\newcommand{\snrmin}{\text{SNR}_{\text{min}}}
\newcommand{\snrmax}{\text{SNR}_{\text{max}}}
\def\rvW{{\mathbf{W}}}
\begin{document}

{\let\thefootnote\relax\footnote{* Equal contribution.}}
\maketitle

\begin{abstract}
Diffusion-based generative models have demonstrated a capacity for perceptually impressive synthesis, but can they also be great likelihood-based models? We answer this in the affirmative, and introduce a family of diffusion-based generative models that obtain state-of-the-art likelihoods on standard image density estimation benchmarks. Unlike other diffusion-based models, our method allows for efficient optimization of the noise schedule jointly with the rest of the model. We show that the variational lower bound (VLB) simplifies to a remarkably short expression in terms of the signal-to-noise ratio of the diffused data, thereby improving our theoretical understanding of this model class. Using this insight, we prove an equivalence between several models proposed in the literature. In addition, we show that the continuous-time VLB is invariant to the noise schedule, except for the signal-to-noise ratio at its endpoints. This enables us to learn a noise schedule that minimizes the variance of the resulting VLB estimator, leading to faster optimization. Combining these advances with architectural improvements, we obtain state-of-the-art likelihoods on image density estimation benchmarks, outperforming autoregressive models that have dominated these benchmarks for many years, with often significantly faster optimization. In addition, we show how to use the model as part of a bits-back compression scheme, and demonstrate lossless compression rates close to the theoretical optimum. Code is available at \url{https://github.com/google-research/vdm}.

\end{abstract}

\section{Introduction}
\begin{figure}[bht]
\begin{subfigure}{.485\textwidth}
  \centering
  \includegraphics[width=.99\linewidth]{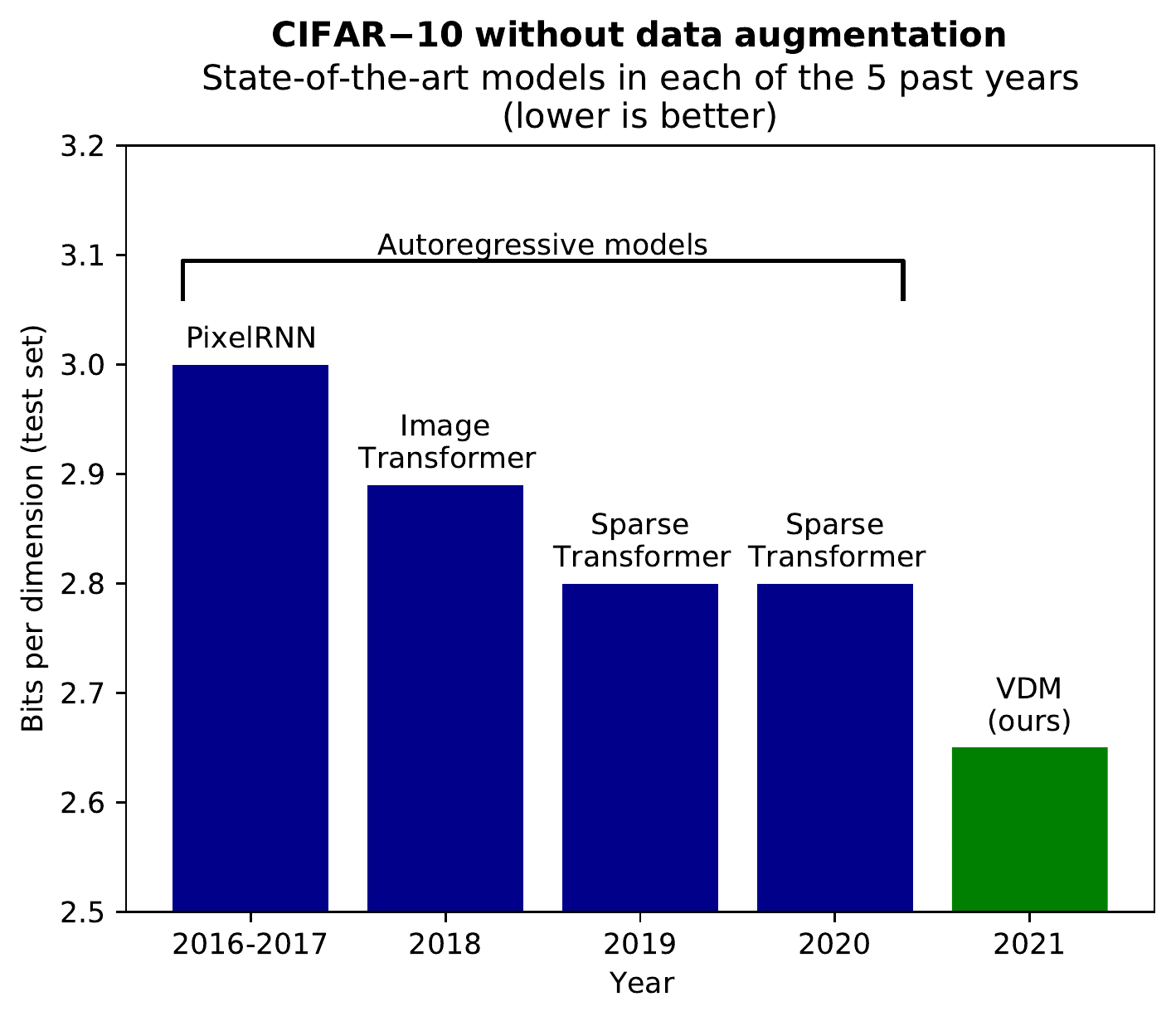}
  \caption{CIFAR-10 without data augmentation}
\end{subfigure}
\begin{subfigure}{.495\textwidth}
  \centering
  \includegraphics[width=.99\linewidth]{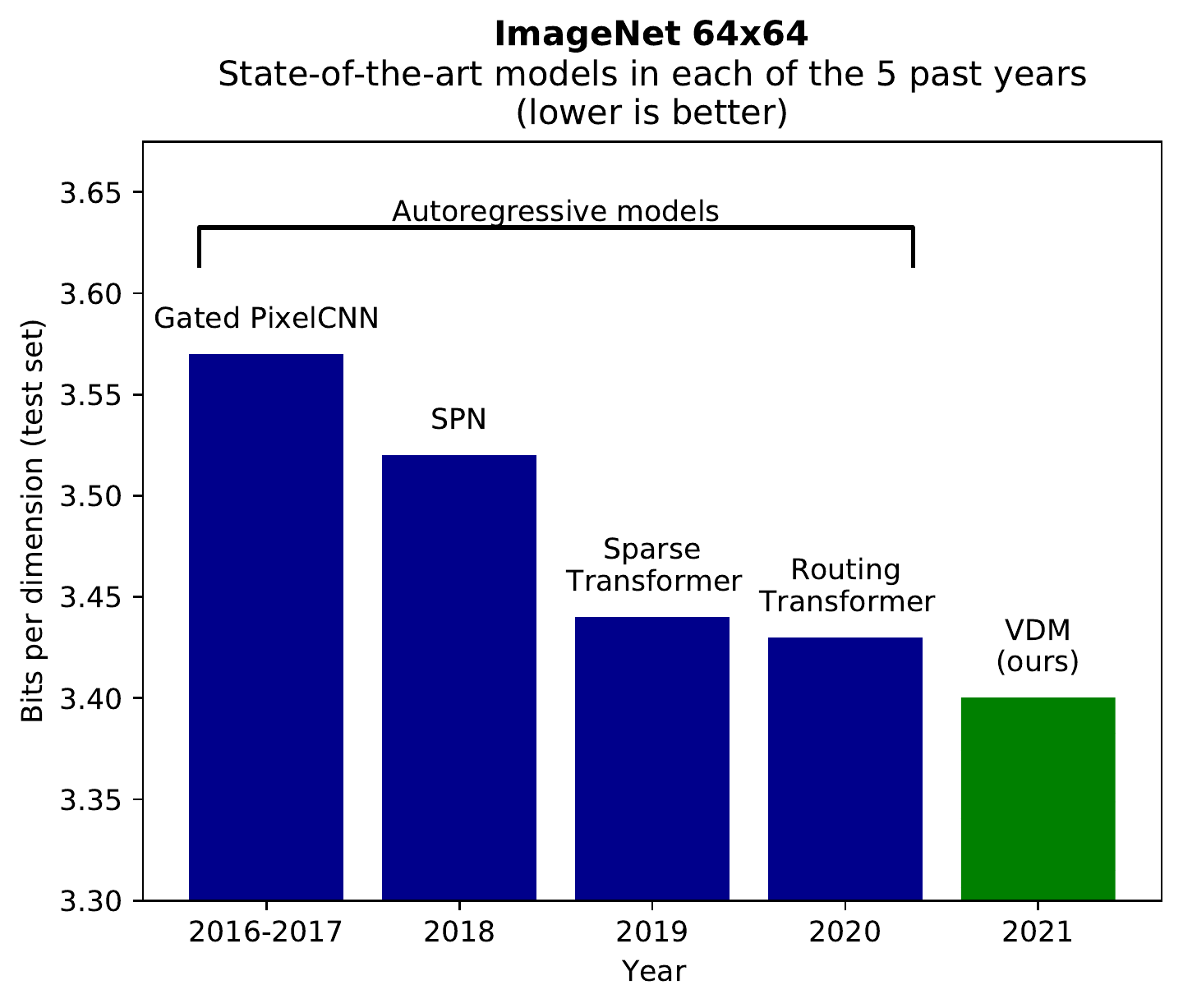}  
  \caption{ImageNet 64x64}
\end{subfigure}
\caption{Autoregressive generative models were long dominant in standard image density estimation benchmarks. In contrast, we propose a family of diffusion-based generative models, \emph{Variational Diffusion Models} (VDMs), that outperforms contemporary autoregressive models in these benchmarks. See Table \ref{table:results} for more results and comparisons.}
\label{fig:sota}
\end{figure}

Likelihood-based generative modeling is a central task in machine learning that is the basis for a wide range of applications ranging from speech synthesis~\citep{oord2016wavenet}, to translation~\citep{sutskever2014sequence}, to compression~\citep{mackay2003information}, to many others. Autoregressive models have long been the dominant model class on this task due to their tractable likelihood and expressivity, as shown in Figure~\ref{fig:sota}. Diffusion models have recently shown impressive results in image \citep{ho2020denoising, song2020score,nichol2021improved} and audio generation \citep{kong2020diffwave, chen2020wavegrad} in terms of perceptual quality, but have yet to match autoregressive models on density estimation benchmarks. In this paper we make several technical contributions that allow diffusion models to challenge the dominance of autoregressive models in this domain.
Our main contributions are as follows:
\begin{itemize}
\item We introduce a flexible family of diffusion-based generative models that achieve new state-of-the-art log-likelihoods on standard image density estimation benchmarks (CIFAR-10 and ImageNet). This is enabled by incorporating Fourier features into the diffusion model and using a learnable specification of the diffusion process, among other modeling innovations.
\item We improve our theoretical understanding of density modeling using diffusion models by analyzing their variational lower bound (VLB), deriving a remarkably simple expression in terms of the signal-to-noise ratio of the diffusion process. This result delivers new insight into the model class: for the continuous-time (infinite-depth) setting we prove a novel invariance of the generative model and its VLB to the specification of the diffusion process, and we show that various diffusion models from the literature are equivalent up to a trivial time-dependent rescaling of the data.
\end{itemize}

\section{Related work}
\label{sec:relatedwork}

Our work builds on diffusion probabilistic models (DPMs)~\citep{sohl2015deep}, or \emph{diffusion models} in short. DPMs can be viewed as a type of variational autoencoder (VAE)~\citep{kingma2013auto,rezende2014stochastic}, whose structure and loss function allows for efficient training of arbitrarily deep models. Interest in diffusion models has recently reignited due to their impressive image generation results \citep{ho2020denoising, song2020improved}.

\citet{ho2020denoising} introduced a number of model innovations to the original DPM, with impressive results on image generation quality benchmarks. They showed that the VLB objective, for a diffusion model with discrete time and diffusion variances shared across input dimensions, is equivalent to multi-scale denoising score matching, up to particular weightings per noise scale. Further improvements were proposed by \citet{nichol2021improved}, resulting in better log-likelihood scores. \citet{gao2020learning} show how diffusion can also be used to efficiently optimize energy-based models (EBMs) towards a close approximation of the log-likelihood objective, resulting in high-fidelity samples even after long MCMC chains.

\citet{song2019generative} first proposed learning generative models through a multi-scale denoising score matching objective, with improved methods in \citet{song2020improved}. This was later extended to continuous-time diffusion with novel sampling algorithms based on reversing the diffusion process~\citep{song2020score}.

Concurrent to our work, \cite{song2021maximum}, \cite{huang2021variational}, and \cite{vahdat2021score} also derived variational lower bounds to the data likelihood under a continuous-time diffusion model. Where we consider the infinitely deep limit of a standard VAE, \cite{song2021maximum} and \cite{vahdat2021score} present different derivations based on stochastic differential equations. \cite{huang2021variational} considers both perspectives and discusses the similarities between the two approaches. An advantage of our analysis compared to these other works is that we present an intuitive expression of the VLB in terms of the signal-to-noise ratio of the diffused data, leading to much simplified expressions of the discrete-time and continuous-time loss, allowing for simple and numerically stable implementation. This also leads to new results on the invariance of the generative model and its VLB to the specification of the diffusion process. We empirically compare to these works, as well as others, in Table~\ref{table:results}.

Previous approaches to diffusion probabilistic models fixed the diffusion process; in contrast optimize the diffusion process parameters jointly with the rest of the model. This turns the model into a type of VAE~\citep{kingma2013auto,rezende2014stochastic}. This is enabled by directly parameterizing the mean and variance of the marginal $q(\rvz_t|\rvz_0)$, where previous approaches instead parameterized the individual diffusion steps $q(\rvz_{t+\epsilon}|\rvz_t)$. In addition, our denoising models include several architecture changes, the most important of which is the use of Fourier features, which enable us to reach much better likelihoods than previous diffusion probabilistic models.

\section{Model}
\label{sec:model}
We will focus on the most basic case of generative modeling, where we have a dataset of observations of $\rvx$, and the task is to estimate the marginal distribution $p(\rvx)$. As with most generative models, the described methods can be extended to the case of multiple observed variables, and/or the task of estimating conditional densities $p(\rvx|\rvy)$. The proposed latent-variable model consists of a diffusion process (Section ~\ref{sec:inferencemodel}) that we invert to obtain a hierarchical generative model (Section ~\ref{sec:generativemodel}). As we will show, the model choices below result in a surprisingly simple variational lower bound (VLB) of the marginal likelihood, which we use for optimization of the parameters.

\subsection{Forward time diffusion process}\label{sec:inferencemodel}
Our starting point is a Gaussian diffusion process that begins with the data $\rvx$, and defines a sequence of increasingly noisy versions of $\rvx$ which we call the \emph{latent variables} $\rvz_t$, where $t$ runs from $t=0$ (least noisy) to $t=1$ (most noisy).
The distribution of latent variable $\rvz_t$ conditioned on $\rvx$, for any $t \in [0,1]$ is given by:
\begin{align}
q(\rvz_t|\rvx) &= \mathcal{N}\left(\alpha_t \rvx, \sigma^2_t \bfI\right),
\label{eq:zt_given_x}
\end{align}
where  $\alpha_t$ and $\sigma^{2}_t$ are strictly positive scalar-valued functions of $t$. Furthermore, let us define the \emph{signal-to-noise ratio} (SNR): \begin{align}
    \snr(t) = \alpha^2_t / \sigma^2_t.
\end{align}
We assume that the $\snr(t)$ is strictly monotonically decreasing in time, i.e. that $\snr(t) < \snr(s)$ for any $t > s$. This formalizes the notion that the $\rvz_t$ is increasingly noisy as we go forward in time. We also assume that both $\alpha_{t}$ and $\sigma^{2}_t$ are smooth, such that their derivatives with respect to time $t$ are finite.  This diffusion process specification includes the \emph{variance-preserving} diffusion process as used by ~\citep{sohl2015deep,ho2020denoising} as a special case, where $\alpha_t = \sqrt{1 - \sigma^2_t}$. Another special case is the \emph{variance-exploding} diffusion process as used by \citep{song2019generative,song2020score}, where $\alpha^2_t = 1$. In experiments, we use the variance-preserving version.

The distributions $q(\rvz_t|\rvz_s)$ for any $t > s$ are also Gaussian, and given in Appendix \ref{sec:distributions}. The joint distribution of latent variables $(\rvz_s, \rvz_t, \rvz_u)$ at any subsequent timesteps $0 \leq s < t < u \leq 1$ is Markov: $q(\rvz_u | \rvz_t, \rvz_s) = q(\rvz_u | \rvz_t)$. Given the distributions above, it is relatively straightforward to verify through Bayes rule that $q(\rvz_s|\rvz_t,\rvx)$, for any $0 \leq s < t \leq 1$, is also Gaussian. This distribution is also given in Appendix \ref{sec:distributions}. 

\subsection{Noise schedule}
\label{sec:learning_noise}

In previous work, the noise schedule has a fixed form (see \cref{app:monotonicnn}, \cref{fig:snr_schedule}). In contrast, we learn this schedule through the parameterization
\begin{align}
\sigma^{2}_t = \text{sigmoid}(\gamma_{\boldeta}(t))
\end{align}
where $\gamma_{\boldeta}(t)$ is a monotonic neural network with parameters ${\boldeta}$, as detailed in Appendix \ref{app:monotonicnn}.

Motivated by the equivalence discussed in Section \ref{sec:schedule_invariance}, we use $\alpha_t = \sqrt{1 - \sigma^2_t}$ in our experiments for both the discrete-time and continuous-time models, i.e. variance-preserving diffusion processes. It is straightforward to verify that $\alpha^2_t$ and $\snr(t)$, as a function of $\gamma_{\boldeta}(t)$, then simplify to:
\begin{align}
\alpha^{2}_t = \text{sigmoid}(-\gamma_{\boldeta}(t)) \\
\snr(t) = \exp(-\gamma_{\boldeta}(t))
\end{align}

\subsection{Reverse time generative model}\label{sec:generativemodel}
We define our generative model by inverting the diffusion process of Section~\ref{sec:inferencemodel}, yielding a hierarchical generative model that samples a sequence of latents $\rvz_t$, with time running backward from $t=1$ to $t=0$.
We consider both the case where this sequence consists of a finite number of steps $T$, as well as a continuous time model corresponding to $T\rightarrow\infty$. We start by presenting the discrete-time case. 

Given finite $T$, we discretize time uniformly into $T$ timesteps (segments) of width $\tau = 1/T$. Defining $s(i) = (i-1)/T$ and $t(i) = i/T$, our hierarchical generative model for data $\rvx$ is then given by:
\begin{align}
p(\rvx) &= \int_{\rvz} p(\rvz_{1})p(\rvx|\rvz_{0}) \prod_{i=1}^T p(\rvz_{s(i)}|\rvz_{t(i)}).
\end{align}
With the variance preserving diffusion specification and sufficiently small $\snr(1)$, we have that $q(\rvz_1|\rvx) \approx \mathcal{N}(\rvz_1; 0,\bfI)$. We therefore model the marginal distribution of $\rvz_1$ as a spherical Gaussian: 
\begin{align}
    p(\rvz_1) = \mathcal{N}(\rvz_1; 0, \bfI).
\end{align}
We wish to choose a model $p(\rvx|\rvz_0)$ that is close to the unknown $q(\rvx|\rvz_0)$. Let $x_i$ and $z_{0,i}$ be the $i$-th elements of $\rvx,  \rvz_0$, respectively. We then use a factorized distribution of the form:
\begin{align}
p(\rvx|\rvz_0) = \prod_i p(x_i|z_{0,i})\label{eq:recon},
\end{align}
where we choose $p(x_i |z_{0,i}) \propto q(z_{0,i}|x_i)$, which is normalized by summing over all possible discrete values of $x_i$ (256 in the case of 8-bit image data). With sufficiently large $\snr(0)$, this becomes a very close approximation to the true $q(\rvx|\rvz_0)$, as the influence of the unknown data distribution $q(\rvx)$ is overwhelmed by the likelihood $q(\rvz_0|\rvx)$.
Finally, we choose the conditional model distributions as \begin{align}p(\rvz_s|\rvz_t) = q(\rvz_s|\rvz_t, \rvx=\hat{\rvx}_{\bT}(\rvz_t; t)),
\end{align}
i.e.\ the same as $q(\rvz_s|\rvz_t,\mathbf{x})$, but with the original data $\rvx$ replaced by the output of a \emph{denoising model} $\hat{\rvx}_{\bT}(\rvz_t; t)$ that predicts $\rvx$ from its noisy version $\rvz_t$. Note that in practice we parameterize the denoising model as a function of a \emph{noise prediction model} (Section \ref{sec:noisepred}), bridging the gap with previous work on diffusion models \citep{ho2020denoising}. The means and variances of $p(\rvz_s|\rvz_t)$ simplify to a remarkable degree; see Appendix \ref{sec:distributions}. 

\subsection{Noise prediction model and Fourier features}
\label{sec:noisepred}

We parameterize the denoising model in terms of a \emph{noise prediction model} $\hat{\bepsilon}_{\bT}(\rvz_t; t)$:
\begin{align}
    \hat{\rvx}_{\bT}(\rvz_t; t) = (\rvz_t - \sigma_t \hat{\bepsilon}_{\bT}(\rvz_t; t))/\alpha_t,
\label{eq:noisepred}
\end{align}
where $\hat{\bepsilon}_{\bT}(\rvz_t; t)$ is parameterized as a neural network. The noise prediction models we use in experiments closely follow~\cite{ho2020denoising}, except that they process the data solely at the original resolution. The exact parameterization of the noise prediction model and noise schedule is discussed in Appendix \ref{sec:hps}.

Prior work on diffusion models has mainly focused on the perceptual quality of generated samples, which emphasizes coarse scale patterns and global consistency of generated images. Here, we optimize for likelihood, which is sensitive to fine scale details and exact values of individual pixels.  To capture the fine scale details of the data, we propose adding a set of \emph{Fourier features} to the input of our noise prediction model. Let $\rvx$ be the original data, scaled to the range $[-1,1]$, and let $\rvz$ be the resulting latent variable, with similar magnitudes. We then append channels $\sin(2^n\pi\rvz)$ and $\cos(2^n\pi\rvz)$, where $n$ runs over a range of integers $\{n_{min}, ..., n_{max}\}$. These features are high frequency periodic functions that amplify small changes in the input data~$\rvz_t$; see Appendix \ref{sec:fourier} for further details.
Including these features in the input of our denoising model leads to large improvements in likelihood as demonstrated in Section~\ref{sec:experiments} and Figure \ref{fig:fourier-ablation}, especially when combined with a learnable SNR function. We did not observe such improvements when incorporating Fourier features into autoregressive models.

\subsection{Variational lower bound}
We optimize the parameters towards the variational lower bound (VLB) of the marginal likelihood, which is given by
\begin{align}
    - \log p(\rvx) \leq %
    - \text{VLB}(\rvx) =
    \underbrace{%
    D_{KL}(q(\rvz_1|\rvx)||p(\rvz_1))
    }_{\text{Prior loss}}
    + \underbrace{%
    \E_{q(\rvz_0|\rvx)}\left[ - \log p(\rvx|\rvz_0) \right]
    }_{\text{Reconstruction loss}}
    +
    \underbrace{
    \lT.
    }_{\text{Diffusion loss}}
    \label{eq:finite_elbo_1}
\end{align}
The prior loss and reconstruction loss can be (stochastically and differentiably) estimated using standard techniques; see \citep{kingma2013auto}. The diffusion loss, $\lT$, is more complicated, and depends on the hyperparameter $T$ that determines the depth of the generative model.

\section{Discrete-time model}

In the case of finite $T$, using $s(i) = (i-1)/T$, $t(i) = i/T$, the diffusion loss is:
\begin{align}
    \lT
    &= 
    \sum_{i=1}^T
    \E_{q(\rvz_{t(i)}|\rvx)} 
    D_{KL}[q(\rvz_{s(i)}|\rvz_{t(i)},\rvx)||p(\rvz_{s(i)}|\rvz_{t(i)})].
    \label{eq:main_finite_ELBO}
\end{align}
In appendix \ref{sec:objective} we show that this expression simplifies considerably, yielding:
\begin{align}
\lT =
\frac{T}{2}
\E_{\bepsilon \sim \mathcal{N}(0,\bfI), i \sim U\{1, T\}}
\left[
\left(\text{SNR}(s)-\text{SNR}(t)\right)
||\rvx - \hat{\rvx}_{\bT}(\rvz_t;t) ||_2^2
\right],
\label{eq:lT_main}\end{align}
where $U\{1, T\}$ is the uniform distribution on the integers $\{1,\ldots,T\}$, and $\rvz_t = \alpha_t \rvx + \sigma_t \bepsilon$. This is the general discrete-time loss for any choice of forward diffusion parameters $(\sigma_t,  \alpha_t)$.  When plugging in the specifications of $\sigma_t$, $\alpha_t$ and $\hat{\rvx}_{\bT}(\rvz_t; t)$ that we use in experiments, given in Sections \ref{sec:learning_noise} and \ref{sec:noisepred}, the loss simplifies to:
\begin{align}
\lT =
\frac{T}{2}
\E_{\bepsilon \sim \mathcal{N}(0,\bfI), i \sim U\{1, T\}}
\left[
(\exp(\gamma_{\boldeta}(t)-\gamma_{\boldeta}(s))-1)
\left\lVert\bepsilon - \hat{\bepsilon}_{\bT}(\rvz_t; t) \right\lVert_2^2
\right]
\label{eq:discrete}\end{align}
where $\rvz_t = \text{sigmoid}(-\gamma_{\boldeta}(t)) \rvx + \text{sigmoid}(\gamma_{\boldeta}(t)) \bepsilon$. 
In the discrete-time case, we simply jointly optimize $\boldeta$ and $\bT$ by maximizing the VLB through a Monte Carlo estimator of \Eqref{eq:discrete}.

Note that $\exp(.)-1$ has a numerically stable primitive $expm1(.)$ in common numerical computing packages; see figure \ref{fig:expm1}. \Eqref{eq:discrete} allows for numerically stable implementation in 32-bit or lower-precision floating point, in contrast with previous implementations of discrete-time diffusion models (e.g. ~\citep{ho2020denoising}), which had to resort to 64-bit floating point.

\subsection{More steps leads to a lower loss}
\label{sec:moresteps}
\begin{figure}[t]
\includegraphics[width=0.9\textwidth]{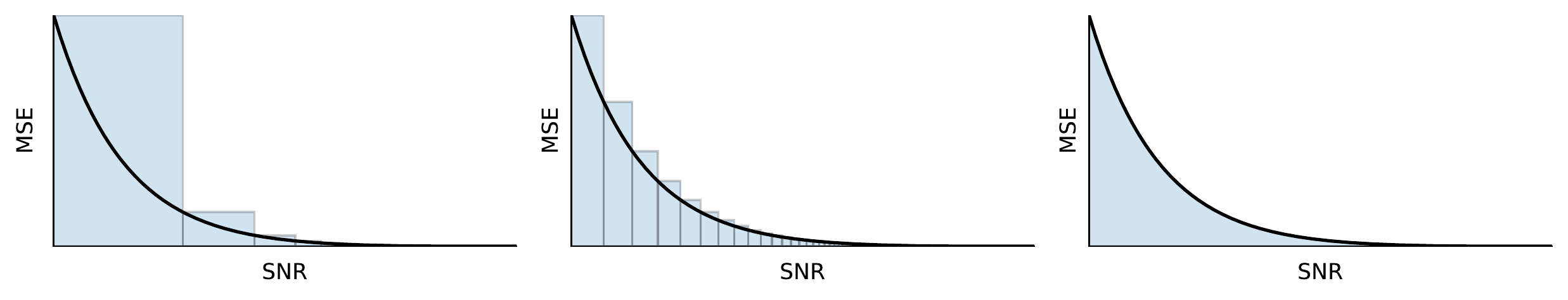}
  \caption{
Illustration of the diffusion loss with few segments $T$ (left), more segments $T$ (middle), and infinite segments $T$ (continuous time, right). The continuous-time loss (\Eqref{eq:elbo_v}) is an integral of mean squared error (MSE) over SNR, here visualized as a black curve. The black curve is strictly decreasing when the model is sufficiently well trained, so the discrete-time loss (\Eqref{eq:lT_main}) is an upper bound (an upper Riemann sum approximation) of this integral that becomes better when segments are added.}
  \label{fig:more_steps_is_better}
\end{figure}

A natural question to ask is what the number of timesteps $T$ should be, and whether more timesteps is always better in terms of the VLB. In Appendix~\ref{appendix:more_steps} we analyze the difference between the diffusion loss with $T$ timesteps, $\lT$, and the diffusion loss with double the timesteps, $\ldT$, while keeping the SNR function fixed. 
We then find that if our trained denoising model $\hat{\rvx}_{\bT}$ is sufficiently good, we have that $\ldT < \lT$, i.e.\ that our VLB will be better for a larger number of timesteps. Intuitively, the discrete time diffusion loss is an upper Riemann sum  approximation of an integral of a strictly decreasing function, meaning that a finer approximation yields a lower diffusion loss. This result is illustrated in Figure~\ref{fig:more_steps_is_better}.

\section{Continuous-time model: $T\rightarrow\infty$}

Since taking more time steps leads to a better VLB, we now take $T\rightarrow\infty$, effectively treating time $t$ as continuous rather than discrete. The model for $p(\rvz_t)$ can in this case be described as a continuous time diffusion process \citep{song2020score} governed by a stochastic differential equation; see Appendix \ref{sec:sde}. In Appendix \ref{sec:objective} we show that in this limit the diffusion loss $\lT$ simplifies further. Letting $\snr'(t) = d \snr(t)/dt$, we have, with $\rvz_t = \alpha_t \rvx + \sigma_t \bepsilon$:
\begin{align}
\linfty
&= -\frac{1}{2}\E_{\bepsilon\sim\mathcal{N}(0,\bfI)} \int_{0}^{1} \snr'(t) \left\rVert \rvx - \hat{\rvx}_{\bT}(\rvz_t;t) \right\lVert_{2}^{2} dt, \label{eq:fusion_integral}
\\
&= -\frac{1}{2}\E_{\bepsilon\sim\mathcal{N}(0,\bfI), t \sim \mathcal{U}(0,1)}
\left[ \snr'(t) \left\rVert \rvx - \hat{\rvx}_{\bT}(\rvz_t;t) \right\lVert_{2}^{2} \right].
\label{eq:fusion}\end{align}
This is the general continuous-time loss for any choice of forward diffusion parameters $(\sigma_t,  \alpha_t)$. When plugging in the specifications of $\sigma_t$, $\alpha_t$ and $\hat{\rvx}_{\bT}(\rvz_t; t)$ that we use in experiments, given in Sections \ref{sec:learning_noise} and \ref{sec:noisepred}, the loss simplifies to:
\begin{align}
\linfty
&= \frac{1}{2}\E_{\bepsilon\sim\mathcal{N}(0,\bfI), t \sim \mathcal{U}(0,1)}
\left[ \gamma'_{\boldeta}(t) \left\rVert \bepsilon - \hat{\bepsilon}_{\bT}(\rvz_t;t) \right\lVert_{2}^{2} \right],
\end{align}
where $\gamma'_{\boldeta}(t) = d \gamma_{\boldeta}(t)/dt$. We use the Monte Carlo estimator of this loss for evaluation and optimization.

\subsection{Equivalence of diffusion models in continuous time}
\label{sec:schedule_invariance}
The signal-to-noise function $\snr(t)$ is invertible due to the monotonicity assumption in Section \ref{sec:inferencemodel}. Due to this invertibility, we can perform a change of variables, and make everything a function of $v \equiv \snr(t)$ instead of $t$, such that $t = \snr^{-1}(v)$. Let $\alpha_v$ and $\sigma_v$ be the functions $\alpha_t$ and $\sigma_t$ evaluated at $t = \snr^{-1}(v)$, and correspondingly let $\rvz_v = \alpha_v \rvx + \sigma_v \bepsilon$. Similarly, we rewrite our noise prediction model as $\tilde{\rvx}_{\bT}(\rvz, v) \equiv \hat{\rvx}_{\bT}(\rvz, \snr^{-1}(v))$. With this change of variables, our continuous-time loss in \Eqref{eq:fusion_integral} can equivalently be written as:
\begin{align}
\linfty = \frac{1}{2}\E_{\bepsilon\sim\mathcal{N}(0,\bfI)} \int_{\snrmin}^{\snrmax} \left\rVert \rvx - \tilde{\rvx}_{\bT}(\rvz_v, v) \right\lVert_{2}^{2} dv,
\label{eq:elbo_v}
\end{align}
where instead of integrating w.r.t. time $t$ we now integrate w.r.t. the signal-to-noise ratio $v$, and where $\snr_\text{min}=\snr(1)$ and $\snr_\text{max}=\snr(0)$.

What this equation shows us is that the only effect the functions $\alpha(t)$ and $\sigma(t)$ have on the diffusion loss is through the values $\snr(t) = \alpha^{2}_t/\sigma^{2}_t$ at endpoints $t=0$ and $t=1$. Given these values $\snr_\text{max}$ and $\snr_\text{min}$, the diffusion loss is invariant to the shape of function $\snr(t)$ between $t=0$ and $t=1$. The VLB is thus only impacted by the function $\snr(t)$ through its endpoints $\snrmin$ and $\snrmax$. 

Furthermore, we find that the distribution $p(\rvx)$ defined by our generative model is also invariant to the specification of the diffusion process. Specifically, let $p^{A}(\rvx)$ denote the distribution defined by the combination of a diffusion specification and denoising function $\{\alpha^A_v, \sigma^A_v, \tilde{\rvx}^A_{\bT}\}$, and similarly let $p^{B}(\rvx)$ be the distribution defined through a different specification $\{\alpha^B_v, \sigma^B_v, \tilde{\rvx}^B_{\bT}\}$, where both specifications have equal $\snr_\text{min}, \snr_\text{max}$; as shown in Appendix~\ref{sec:app_equivalence}, we then have that $p^{A}(\rvx) = p^{B}(\rvx)$ if $\tilde{\rvx}^B_{\bT}(\rvz, v) \equiv \tilde{\rvx}^A_{\bT}((\alpha^A_v/\alpha^B_v)\rvz, v)$. The distribution on all latents $\rvz_v$ is then also the same under both specifications, up to a trivial rescaling. This means that any two diffusion models satisfying the mild constraints set in 3.1 (which includes e.g.\ the \emph{variance exploding} and \emph{variance preserving} specifications considered by~\cite{song2020score}), can thus be seen as equivalent in continuous time.

\subsection{Weighted diffusion loss}
This equivalence between diffusion specifications continues to hold even if, instead of the VLB, these models optimize a \emph{weighted} diffusion loss of the form:
\begin{align}
\mathcal{L}_{\infty}(\rvx, w) = \frac{1}{2}\E_{\bepsilon\sim\mathcal{N}(0,\bfI)} \int_{\snr_{\text{min}}}^{\snr_{\text{max}}} w(v) \left\rVert \rvx - \tilde{\rvx}_{\bT}(\rvz_v, v) \right\lVert_{2}^{2} dv,
\label{eq:weighted_diff}
\end{align}
which e.g.\ captures all the different objectives discussed by \cite{song2020score}, see Appendix~\ref{sec:ncsn_ddpm_weights}. Here, $w(v)$ is a weighting function that generally puts increased emphasis on the noisier data compared to the VLB, and which thereby can sometimes improve perceptual generation quality as measured by certain metrics like FID and Inception Score. For the models presented in this paper, we further use $w(v)=1$ as corresponding to the (unweighted) VLB.

\subsection{Variance minimization}

Lowering the variance of the Monte Carlo estimator of the continuous-time loss generally improves the efficiency of optimization. We found that using a low-discrepancy sampler for $t$, as explained in Appendix \ref{sec:lowdisc}, leads to a significant reduction in variance. In addition, due to the invariance shown in Section~\ref{sec:schedule_invariance} for the continous-time case, we can optimize the schedule \emph{between} its endpoints w.r.t. to minimize the variance of our estimator of loss, as detailed in Appendix \ref{app:variance_minimization}. The endpoints of the noise schedule are simply optimized w.r.t. the VLB. 

\section{Experiments}
\label{sec:experiments}

\begin{table}[t]
\footnotesize
\begin{center}
\setlength{\tabcolsep}{4pt}
\begin{tabular}{lccccc}
\textbf{Model} & Type & CIFAR10 & CIFAR10 & ImageNet & ImageNet %
\\
(Bits per dim on test set)&  & no data aug. & data aug. & 32x32 & 64x64%
\\
\midrule
\textit{Previous work}\\
ResNet VAE with IAF~\citep{kingma2016improving} & VAE & 3.11 & \\
Very Deep VAE~\citep{child2020very} & VAE & 2.87 & & 3.80 & 3.52 \\
NVAE~\citep{vahdat2020nvae} & VAE & 2.91 & & 3.92 & \\
Glow~\citep{kingma2018glow} & Flow & & $3.35^{(B)}$ & 4.09 & 3.81 \\
Flow++~\citep{ho2019flow++} & Flow & 3.08 & & 3.86 & 3.69 \\
PixelCNN~\citep{van2016pixel} & AR & 3.03 & & 3.83 & 3.57\\
PixelCNN++~\citep{salimans2017pixelcnn++} & AR & 2.92 &\\
Image Transformer~\citep{parmar2018image} & AR & 2.90 & & 3.77\\
SPN ~\citep{menick2018generating} & AR & & & & 3.52 \\
Sparse Transformer~\citep{child2019generating}& AR &2.80 & & & 3.44 %
\\
Routing Transformer~\citep{roy2021efficient} & AR & & & & 3.43\\
Sparse Transformer + DistAug~\citep{jun2020distribution} & AR & & $2.53^{(A)}$ & & \\
DDPM~\citep{ho2020denoising} & Diff & & $3.69^{(C)}$ & & %
\\
EBM-DRL~\citep{gao2020learning} & Diff & & $3.18^{(C)}$ & & \\
Score SDE~\citep{song2020score} & Diff & 2.99 & & & %
\\
Improved DDPM~\citep{nichol2021improved} & Diff & 2.94 & & & 3.54 %
\\
\midrule
\textit{Concurrent work}\\
CR-NVAE \citep{sinha2021consistency} & VAE & & $2.51^{(A)}$ \\ 
LSGM~\citep{vahdat2021score} & Diff & 2.87\\
ScoreFlow~\citep{song2021maximum} (variational bound) & Diff & & $2.90^{(C)}$ & 3.86 & \\
ScoreFlow~\citep{song2021maximum} (cont. norm. flow) & Diff & 2.83 & $2.80^{(C)}$ & 3.76 & \\
\midrule
\textit{Our work}\\
\textbf{VDM (variational bound)} & Diff & \textbf{2.65} & $\textbf{2.49}^{(A)}$ & \textbf{3.72} & \textbf{3.40} %
\\
\bottomrule
\end{tabular}
\end{center}
\caption{Summary of our findings for density modeling tasks, in terms of bits per dimension (BPD) on the test set. Model types are autoregressive (AR), normalizing flows (Flow), variational autoencoders (VAE), or diffusion models (Diff). Our results were obtained using the continuous-time formulation of our model. CIFAR-10 data augmentations are: (A) extensive, (B) small translations, or (C) horizontal flips. The numbers for VDM are variational bounds, and can likely be improved by estimating the marginal likelihood through importance sampling, or through evaluation of the corresponding continuous normalizing flow as done by \cite{song2021maximum}.}
\label{table:results}
\end{table}

We demonstrate our proposed class of diffusion models, which we call \emph{Variational Diffusion Models} (VDMs), on the CIFAR-10~\citep{krizhevsky2009learning} dataset, and the downsampled ImageNet~\citep{van2016pixel, deng2009imagenet} dataset, where we focus on maximizing likelihood. For our result with data augmentation we used random flips, 90-degree rotations, and color channel swapping. More details of our model specifications are in Appendix~\ref{sec:hps}.

\subsection{Likelihood and samples}
\begin{figure}[t]
    \centering
    \includegraphics[width=\textwidth]{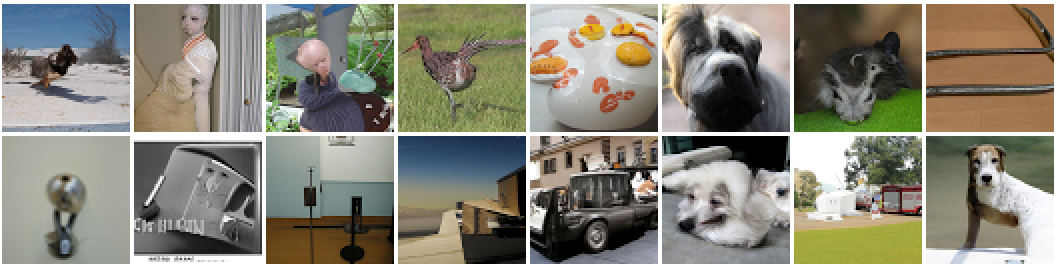}
    \caption{Non cherry-picked unconditional samples from our Imagenet 64x64 model, trained in continuous time and generated using $T=1000$. The model's hyper-parameters and parameters are optimized w.r.t. the likelihood bound, so the model is not optimized for synthesis quality.}
    \label{fig:imagenet64}
\end{figure}

Table~\ref{table:results} shows our results on modeling the CIFAR-10 dataset, and the downsampled ImageNet dataset. We establish a new state-of-the-art in terms of test set likelihood on all considered benchmarks, by a significant margin. Our model for CIFAR-10 without data augmentation surpasses the previous best result of $2.80$ about 10x faster than it takes the Sparse Transformer to reach this, in wall clock time on equivalent hardware.
Our CIFAR-10 model, whose hyper-parameters were tuned for likelihood, results in a FID (perceptual quality) score of 7.41. This would have been state-of-the-art until recently, but is worse than recent diffusion models that specifically target FID scores ~\citep{nichol2021improved,song2020score,ho2020denoising}. By instead using a weighted diffusion loss, with the weighting function $w(\snr)$ used by \cite{ho2020denoising} and described in Appendix~\ref{sec:ncsn_ddpm_weights}, our FID score improves to 4.0. We did not pursue further tuning of the model to improve FID instead of likelihood.
A random sample of generated images from our model is provided in Figure~\ref{fig:imagenet64}. We provide additional samples from this model, as well as our other models for the other datasets, in Appendix~\ref{sec:app_more_samples}.

\subsection{Ablations}
Next, we investigate the relative importance of our contributions. In Table~\ref{table:t_results} we compare our discrete-time and continuous-time specifications of the diffusion model: When evaluating our model with a small number of steps, our discretely trained models perform better by learning the diffusion schedule to optimize the VLB. However, as argued theoretically in Section~\ref{sec:moresteps}, we find experimentally that more steps $T$ indeed gives better likelihood. When $T$ grows large, our continuously trained model performs best, helped by training its diffusion schedule to minimize variance instead. %

\begin{figure}[t]
    \centering\footnotesize
    \begin{subfigure}[b]{0.35\textwidth}
        \centering
        \includegraphics[width=\textwidth]{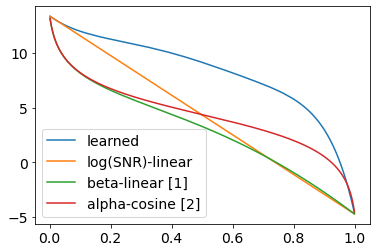}
        \caption{$\log\snr$ vs time $t$ }
        \label{fig:snr_schedule}
    \end{subfigure}\hspace{1.5cm}
    \begin{subfigure}[b]{0.35\textwidth}  
        \centering
        \begin{tabular}{rr}
        $\snr(t)$~schedule & $\mathrm{Var}(\text{BPD})$\\
        \midrule
        \textbf{Learned (ours)} & \textbf{0.53}\\
        $\log\snr$-linear & 6.35\\
        $\beta$-Linear [1] & 31.6\\
        $\alpha$-Cosine [2] & 31.1\\
        \bottomrule
        \end{tabular}
        \caption{Variance of VLB estimate}
        \label{table:variance}
    \end{subfigure}
    \caption{Our learned continuous-time variance-minimizing noise schedule $\snr(t)$ for CIFAR-10, compared to its log-linear initialization and to schedules from the literature: [1] The $\beta$-Linear schedule from \cite{ho2020denoising}, [2] The $\alpha$-Cosine schedule from \cite{nichol2021improved}. All schedules were scaled and shifted on the log scale such that the resulting $\snrmin, \snrmax$ were the equal to our learned endpoints, resulting in the same VLB estimate of 2.66. We report the variance of our VLB estimate per data point, computed on the test set, and conditional on the data: This does not include the noise due to sampling minibatches of data.}
\end{figure}

Minimizing the variance also helps the continuous time model to train faster, as shown in Figure~\ref{fig:fourier-ablation}. This effect is further examined in Table~\ref{table:variance}, where we find dramatic variance reductions compared to our baselines in continuous time. Figure~\ref{fig:snr_schedule} shows how this effect is achieved: Compared to the other schedules, our learned schedule spends much more time in the high $\snr(t)$ / low $\sigma^{2}_{t}$ range.

In Figure~\ref{fig:fourier-ablation} we further show training curves for our model including and excluding the Fourier features proposed in Appendix~\ref{sec:fourier}: with Fourier features enabled our model achieves much better likelihood. For comparison we also implemented Fourier features in a PixelCNN++ model~\citep{salimans2017pixelcnn++}, where we do not see a benefit. In addition, we find that learning the SNR is necessary to get the most out of including Fourier features: if we fix the SNR schedule to that used by \cite{ho2020denoising}, the maximum log-SNR is fixed to approximately 8 (see figure~\ref{fig:weighting_funs}), and test set negative likelihood stays above 4 bits per dim. When learning the SNR endpoints, our maximum log-SNR ends up at $13.3$, which, combined with the inclusion of Fourier features, leads to the SOTA test set likelihoods reported in Table~\ref{table:results}.

\subsection{Lossless compression}
\begin{figure}[t]
\begin{floatrow}
\capbtabbox{%
\scriptsize
\begin{tabular}{rrrc}
$T_{train}$ & $T_{eval}$ & BPD & \makecell{Bits-Back \\ Net BPD} \\
\midrule
10 & 10 & 4.31 \\
100 & 100 & 2.84 \\
250 & 250 & 2.73 \\
500 & 500 & 2.68 \\
1000 & 1000 & 2.67 \\
10000 & 10000 & 2.66 \\
\midrule
$\infty$ & 10 & 7.54 & 7.54 \\
$\infty$ & 100 & 2.90 & 2.91 \\
$\infty$ & 250 & 2.74 & 2.76 \\
$\infty$ & 500 & 2.69 & 2.72 \\
$\infty$ & 1000 & 2.67 & 2.72 \\
$\infty$ & 10000 & \textbf{2.65} \\
\midrule
$\infty$ & $\infty$ & \textbf{2.65}\\
\bottomrule
\end{tabular}
}{%
\caption{Discrete versus continuous-time training and evaluation with CIFAR-10, in terms of bits per dimension (BPD).}%
\label{table:t_results}
}
\ffigbox{%
  \includegraphics[width=0.5\textwidth]{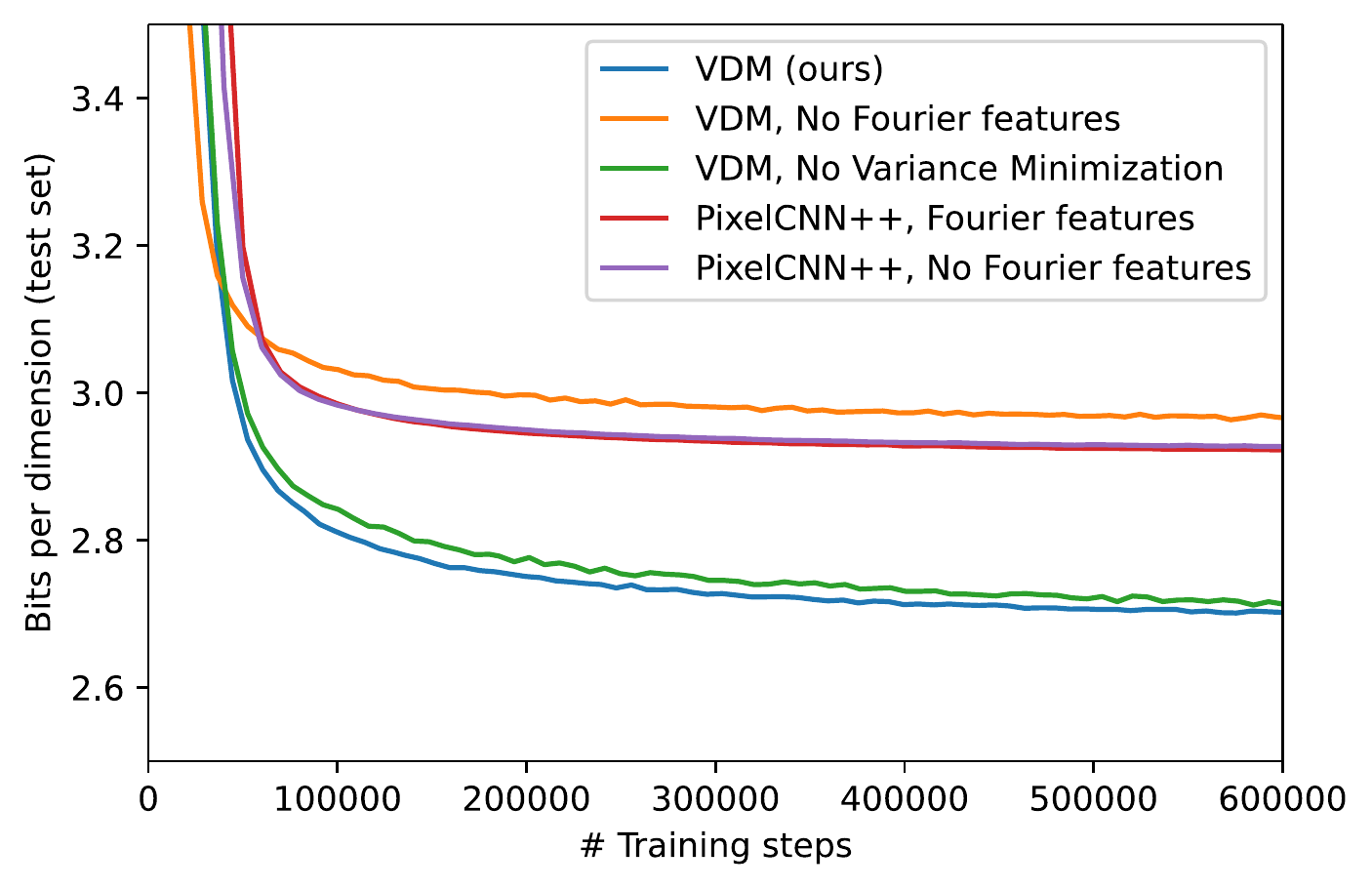}%
}{%
  \caption{Test set likelihoods during training, with/without Fourier features, and with/without learning the noise schedule to minimize variance.}%
  \label{fig:fourier-ablation}
}
\end{floatrow}
\end{figure}

For a fixed number of evaluation timesteps $T_{eval}$, our diffusion model in discrete time is a hierarchical latent variable model that can be turned into a lossless compression algorithm using bits-back coding~\citep{hinton1993keeping}. 
As a proof of concept of practical lossless compression, Table~\ref{table:t_results} reports net codelengths on the CIFAR10 test set for various settings of $T_{eval}$ using BB-ANS~\citep{townsend2018practical}, an implementation of bits-back coding based on asymmetric numeral systems~\citep{duda2009asymmetric}. Details of our implementation are given in Appendix~\ref{sec:compression}. We achieve state-of-the-art net codelengths, proving our model can be used as the basis of a lossless compression algorithm. However, for large $T_{eval}$ a gap remains with the theoretically optimal codelength corresponding to the negative VLB, and compression becomes computationally expensive due to the large number of neural network forward passes required. Closing this gap with more efficient implementations of bits-back coding suitable for very deep models is an interesting avenue for future work.

\section{Conclusion}
\label{sec:conclusion}
We presented state-of-the-art results on modeling the density of natural images using a new class of diffusion models that incorporates a learnable diffusion specification, Fourier features for fine-scale modeling, as well as other architectural innovations. In addition, we obtained new theoretical insight into likelihood-based generative modeling with diffusion models, showing a surprising invariance of the VLB to the forward time diffusion process in continuous time, as well as an equivalence between various diffusion processes from the literature previously thought to be different.

\section*{Acknowledgments}
We thank Yang Song, Kevin Murphy, Mohammad Norouzi and Chin-Yun Yu for helpful feedback on the paper, and Ruiqi Gao for helping with writing an open source version of the code.

\bibliographystyle{plainnat}
\bibliography{references.bib} 

\vfill
\pagebreak
\appendix

\section{Distribution details}
\label{sec:distributions}

\subsection{$q(\rvz_t|\rvz_s)$}
The distribution of $\rvz_t$ given $\rvz_s$, for any $0 \leq s < t \leq 1$, is given by:
\begin{align}
q(\rvz_t|\rvz_s) = \mathcal{N}\left(\alpha_{t|s} \rvz_s, \sigma^2_{t|s} \bfI\right),
\label{eq:zt_given_zs}
\end{align}
where
\begin{align}
\alpha_{t|s} = \alpha_t/\alpha_s,
\end{align}
and
\begin{align}
\sigma_{t|s}^2 = \sigma_t^2 - \alpha_{t|s}^2\sigma_s^2.
\end{align}

\subsection{$q(\rvz_s|\rvz_t,\rvx)$}
Due to the Markov property of the forward process, for $t > s$, we have that $q(\rvz_s, \rvz_t | \rvx) = q(\rvz_s | \rvx) q(\rvz_t | \rvz_s)$.
The term $q(\rvz_s|\rvz_t,\rvx)$ can be viewed as a Bayesian posterior resulting from a prior $q(\rvz_s|\rvx)$, updated with a likelihood term $q(\rvz_t|\rvz_s)$.

Generally, when we have a Gaussian prior of the form $p(x) = \mathcal{N}(\mu_A, \sigma_A^2)$ and a linear-Gaussian likelihood of the form $p(y|x) = \mathcal{N}(a x, \sigma_B^2)$, then the general solution for the posterior is $p(x|y) = \mathcal{N}(\tilde{\mu}, \tilde{\sigma}^2)$, where $\tilde{\sigma}^{-2} = \sigma_A^{-2} + a^2 \sigma_B^{-2}$, and $\tilde{\mu} = \tilde{\sigma}^{-2} (\sigma_A^{-2} \mu_A + a \sigma_B^{-2} y)$.

Plugging our prior term $q(\rvz_s|\rvx) = \mathcal{N}\left(\alpha_s \rvx, \sigma^2_s \bfI\right)$ (\Eqref{eq:zt_given_x}) and linear-Gaussian likelihood term $q(\rvz_t|\rvz_s) = \mathcal{N}(\alpha_{t|s} \rvz_s, \sigma^2_{t|s})$ (\Eqref{eq:zt_given_zs}) into this general posterior equation, yields a posterior:
\begin{align}
q(\rvz_s|\rvz_t,\rvx) &= \mathcal{N}(\bmu_{Q}(\rvz_{t},\rvx; s,t), \sigma^2_Q(s,t) \bfI)\\
\text{where\;\;}
\sigma^{-2}_Q(s,t) 
&= \sigma_s^{-2} + \alpha_{t|s}^{2}\sigma^{-2}_{t|s}\label{eq:postvar}
= \sigma^2_t / (\sigma^2_{t|s}\sigma^2_s)\\
\sigma^{2}_Q(s,t) &= \sigma^2_{t|s}\sigma^2_s / \sigma^2_t\\
\text{and\;\;}
\bmu_{Q}(\rvz_{t},\rvx; s,t) 
&
= \frac{\alpha_{t|s}\sigma^{2}_s}{\sigma^{2}_t}\rvz_{t} + \frac{\alpha_s \sigma^{2}_{t|s}}{\sigma^{2}_{t}}\rvx.
\end{align}

\subsection{$p(\rvz_s|\rvz_t)$}
Finally, we choose the conditional model distributions as
\begin{align}
p(\rvz_s|\rvz_t) = q(\rvz_s|\rvz_t, \rvx=\hat{\rvx}_{\bT}(\rvz_t; t)),
\label{eq:zs_given_zt_model}
\end{align}
i.e.\ the same as $q(\rvz_s|\rvz_t,\mathbf{x})$, but with the original data $\rvx$ replaced by the output of a denoising model $\hat{\rvx}_{\bT}(\rvz_t; t)$ that predicts $\rvx$ from its noisy version $\rvz_t$. We then have
\begin{align}
p(\rvz_s|\rvz_t) = \mathcal{N}(\rvz_s; \bmu_{\bT}(\rvz_{t}; s, t), \sigma^2_Q(s,t) \bfI)
\end{align}
with variance $\sigma^2_Q(s,t)$ the same as in \Eqref{eq:postvar}, and 
\begin{align}
\bmu_{\bT}(\rvz_t; s, t) 
&= \frac{\alpha_{t|s}\sigma^{2}_s}{\sigma^{2}_t}\rvz_{t} + \frac{\alpha_s \sigma^{2}_{t|s}}{\sigma^{2}_{t}}\hat{\rvx}_{\bT}(\rvz_t; t) \nonumber\\
&= \frac{1}{\alpha_{t|s}}\rvz_{t} - \frac{\sigma^{2}_{t|s}}{\alpha_{t|s}\sigma_{t}}\hat{\bepsilon}_{\bT}(\rvz_t; t) \nonumber\\
&= \frac{1}{\alpha_{t|s}}\rvz_t + \frac{\sigma^2_{t|s}}{\alpha_{t|s}} \snT(\rvz_t;t), \label{eq:preds}
\end{align}
where
\begin{align}
\hat{\bepsilon}_{\bT}(\rvz_t; t) = (\rvz_t - \alpha_t\hat{\rvx}_{\bT}(\rvz_t; t))/\sigma_t
\end{align}
and
\begin{align}
\snT(\rvz_t;t) = (\alpha_t\hat{\rvx}_{\bT}(\rvz_t; t) - \rvz_t)/\sigma^{2}_t.
\end{align}

Equation \ref{eq:preds} shows that we can interpret our model in three different ways: 
\begin{enumerate}
\item In terms of the \emph{denoising model} $\hat{\rvx}_{\bT}(\rvz_t; t)$ that recovers $\rvx$ from its corrupted version $\rvz_t$.
\item In terms of a \emph{noise prediction model} $\hat{\bepsilon}_{\bT}(\rvz_t; t)$ that directly infers the noise $\bepsilon$ that was used to generate $\rvz_t$.
\item In terms of a \emph{score model} $\snT(\rvz_t;t)$, that at its optimum equals the scores of the marginal density: $\mathbf{s}^*(\rvz_t;t) = \nabla_{\rvz} \log q(\rvz_t)$; see Appendix \ref{sec:dsm}.
\end{enumerate}
These are three equally valid views on the same model class, that have been used interchangeably in the literature. We find the denoising interpretation the most intuitive, and will therefore mostly use $\hat{\rvx}_{\bT}(\rvz_t; t)$ in the theoretical part of this paper, although in practice we parameterize our model via $\hat{\bepsilon}_{\bT}(\rvz_t; t)$ following \cite{ho2020denoising}. The parameterization of our model is discussed in Appendix \ref{sec:hps}.

\subsection{Further simplification of $p(\rvz_s|\rvz_t)$}
\label{sec:sampling}

\begin{figure}[t]
    \centering
    \includegraphics[width=.8\textwidth]{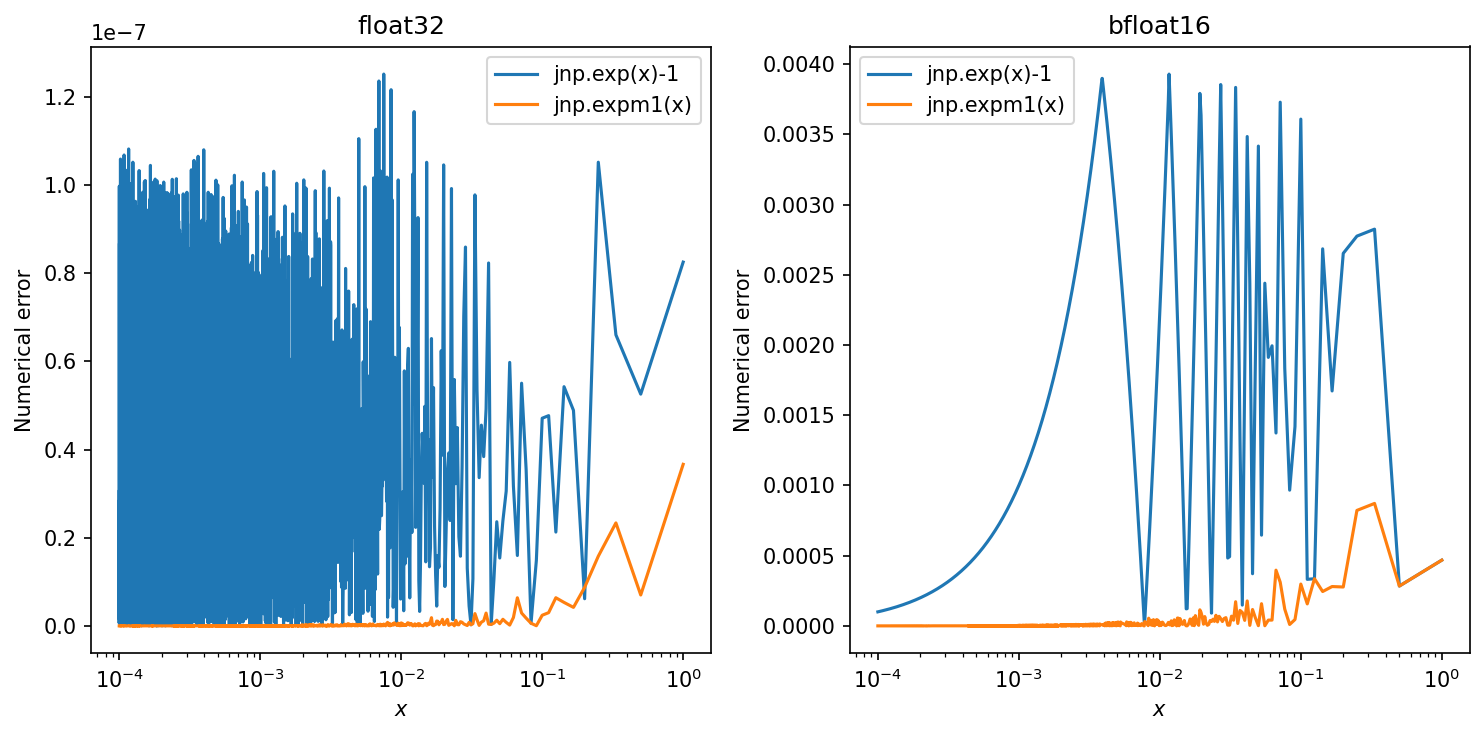}
    \caption{For optimization of the discrete-time diffusion loss and sampling from either the continuous-time or discrete-time model, an operation of the form $\exp(x)-1$ needs to be performed. This operation can result in large numerical errors when performed with 32-bit (float32) or 16-bit (e.g. bfloat16) floating point numbers. For this reason, many numerical packages implement the numerically more stable version $\text{expm1}(x) = \exp(x)-1$. We here plot the absolute numerical errors for each of these versions for float32 and bfloat16. For this plot we used $\text{expm1}(.)$ with 64-bit floating point (float64) as ground truth. The plotted numerical error is the absolute value of the difference between the ground truth and the computed float32 or bfloat16 values.}
    \label{fig:expm1}
\end{figure}

After plugging in the specifications of $\sigma_t$, $\alpha_t$ and $\hat{\rvx}_{\bT}(\rvz_t; t)$ that we use in experiments, given in Sections \ref{sec:learning_noise} and \ref{sec:noisepred}, it can be verified that the distribution $p(\rvz_s|\rvz_t) = \mathcal{N}(\bmu_{\bT}(\rvz_{t}; s, t), \sigma^2_Q(s,t) \bfI)$ simplifies to:
\begin{align}
\bmu_{\bT}(\rvz_{t}; s, t) &= \frac{\alpha_s}{\alpha_t}\left( \rvz_t +
\sigmoid_t
\text{expm1}(\gamma_{\boldeta}(s)-\gamma_{\boldeta}(t)) \hat{\bepsilon}_{\bT}(\rvz_t; t) \right)
\\
\sigma^2_Q(s,t) &= \sigma^2_s \cdot (- \text{expm1}(\gamma_{\boldeta}(s)-\gamma_{\boldeta}(t)))
\end{align}
where $\alpha_s = \sqrt{\text{sigmoid}(-\gamma_{\boldeta}(s))}$, $\alpha_t = \sqrt{\text{sigmoid}(-\gamma_{\boldeta}(t))}$, $\sigmoid^2_s = \text{sigmoid}(\gamma_{\boldeta}(s))$, $\sigmoid_s = \sqrt{\text{sigmoid}(\gamma_{\boldeta}(s))}$, and where $\text{expm1}(.) = \exp(.)-1$; see Figure \ref{fig:expm1}. Ancestral sampling from this distribution can be performed through simply doing:
\begin{align}
    \rvz_s &= \sqrt{\alpha^2_s/\alpha^2_t}(\rvz_t - \sigmoid_t c  \hat{\bepsilon}_{\bT}(\rvz_t; t)) + \sqrt{(1-\alpha^2_s) c} \bepsilon
\end{align}
where $\alpha^2_s = \text{sigmoid}(-\gamma_{\boldeta}(s))$, $\alpha^2_t = \text{sigmoid}(-\gamma_{\boldeta}(t))$, $c = -\text{expm1}(\gamma_{\boldeta}(s) - \gamma_{\boldeta}(t))$, and $\bepsilon \sim \mathcal{N}(0,\bfI)$.

\section{Hyperparameters, architecture, and implementation details}
\label{sec:hps}
In this section we provide details on the exact setup for each of our experiments. In Sections \ref{app:model} we describe the choices in common to each of our experiments. Hyperparameters specific to the individual experiments are given in Section~\ref{app:exp_settings}. We are currently working towards open sourcing our code.

\subsection{Model and implementation}
\label{app:model}
Our denoising models are parameterized in terms of noise prediction models $\hat{\bepsilon}_{\bT}(
\rvz_{t}; \gamma_t)$, as explained in Section \ref{sec:noisepred}. Our noise prediction models $\hat{\bepsilon}_{\bT}$ closely follow the architecture used by \cite{ho2020denoising}, which is based on a U-Net type neural net \citep{ronneberger2015u} that maps from the input $\rvz \in \mathbb{R}^{d}$ to output of the same dimension. As compared to their publically available code at \url{https://github.com/hojonathanho/diffusion}, our implementation differs in the following ways:
\begin{itemize}
    \item Our networks don't perform any internal downsampling or upsampling: we process all the data at the original input resolution.
    \item Our models are deeper than those used by \cite{ho2020denoising}. Specific numbers are given in Section~\ref{app:exp_settings}.
    \item Instead of taking time $t$ as input to the noise prediction model, we use $\gamma_t$, which we rescale to have approximately the same range as $t$ of $[0,1]$ before using it to form 'time' embeddings in the same way as \cite{ho2020denoising}. 
    \item Our models calculate Fourier features on the input data $\rvz_t$ as discussed in Appendix \ref{sec:fourier}, which are then concatenated to $\rvz_t$ before being fed to the U-Net.
    \item Apart from the \emph{middle} attention block that connects the upward and downward branches of the U-Net, we remove all other attention blocks from the model. We found that these attention blocks made it more likely for the model to overfit to the training set.
    \item All of our models use dropout at a rate of 0.1 in the intermediate layers, as did \cite{ho2020denoising}. In addition we regularize the model by using decoupled weight decay \citep{loshchilov2017decoupled} with coefficient 0.01.
    \item We use the Adam optimizer with a learning rate of $2e^{-4}$ and exponential decay rates of $\beta_{1}=0.9, \beta_{2}=0.99$. We found that higher values for $\beta_{2}$ resulted in training instabilities.
    \item For evaluation, we use an exponential moving average of our parameters, calculated with an exponential decay rate of $0.9999$.
\end{itemize}

We regularly evaluate the variational bound on the likelihood on the validation set and find that our models do not overfit during training, using the current settings. We therefore do not use early stopping and instead allow the network to be optimized for 10 million parameter updates for CIFAR-10, and for 2 million updates for ImageNet, before obtaining the test set numbers reported in this paper. It looks like our models keep improving even after this number of updates, in terms of likelihood, but we did not explore this systematically due to resource constraints.

All of our models are trained on TPUv3 hardware (see \url{https://cloud.google.com/tpu}) using data parallelism. We also evaluated our trained models using CPU and GPU to check for robustness of our reported numbers to possible rounding errors. We found only very small differences when evaluating on these other hardware platforms.

\subsection{Settings for each dataset}
\label{app:exp_settings}
Our model for CIFAR-10 with no data augmentation uses a U-Net of depth 32, consisting of 32 ResNet blocks in the forward direction and 32 ResNet blocks in the reverse direction, with a single attention layer and two additional ResNet blocks in the middle. We keep the number of channels constant throughout at 128. This model was trained on 8 TPUv3 chips, with a total batch size of 128 examples. Reaching a test-set BPD of $2.65$ after 10 million updates takes 9 days, although our model already surpasses sparse transformers (the previous state-of-the-art) of $2.80$ BPD after only $2.5$ hours of training.

For CIFAR-10 with data augmentation we used random flips, 90-degree rotations, and color channel swapping, which were previously shown to help for density estimation by~\cite{jun2020distribution}. Each of the three augmentations independently were given a $50\%$ probability of being applied to each example, which means that 1 in 8 training examples was not augmented at all. For this experiment, we doubled the number of channels in our model to 256, and decreased the dropout rate from $10\%$ to $5\%$. Since overfitting was less of a problem with data augmentation, we add back the attention blocks after each ResNet block, following \cite{ho2020denoising}. We also experimented with conditioning our model on an additional binary feature that indicates whether or not the example was augmented, which can be seen as a simplified version of the augmentation conditioning proposed by \cite{jun2020distribution}. Conditioning made almost no difference to our results, which may be explained by the relatively large fraction ($12.5\%$) of clean data fed to our model during training. We trained our model for slightly over a week on 128 TPUv3 chips to obtain the reported result.

Our model for 32x32 ImageNet looks similar to that for CIFAR-10 without data augmentation, with a U-Net depth of 32, but uses double the number of channels at 256. It is trained using data parallelism on 32 TPUv3 chips, with a total batch size of 512.

Our model for 64x64 ImageNet uses double the depth at 64 ResNet layers in both the forward and backward direction in the U-Net. It also uses a constant number of channels of 256. This model is trained on 128 TPUv3 chips at a total batch size of 512 examples. %

\section{Fourier features for improved fine scale prediction}
\label{sec:fourier}
Prior work on diffusion models has mainly focused on the perceptual quality of generated samples, which emphasizes coarse scale patterns and global consistency of generated images. Here, we optimize for likelihood, which is sensitive to fine scale details and exact values of individual pixels. Since our reconstruction model $p(\rvx|\rvz_{0})$ given in \Eqref{eq:recon} is weak, the burden of modeling these fine scale details falls on our denoising diffusion model $\hat{\rvx}_{\bT}$. In initial experiments, we found that the denoising model had a hard time accurately modeling these details. At larger noise levels, the latents $\rvz_t$ follow a smooth distribution due to the added Gaussian noise, but at the smallest noise levels the discrete nature of 8-bit image data leads to sharply peaked marginal distributions $q(\rvz_t)$.%

To capture the fine scale details of the data, we propose adding a set of \emph{Fourier features} to the input of our denoising model $\hat{\rvx}_{\bT}(\rvz_t; t)$. Such Fourier features consist of a linear projection of the original data onto a set of periodic basis functions with high frequency, which allows the network to more easily model high frequency details of the data. Previous work~\citep{tancik2020fourier} has used these features for input coordinates to model high frequency details across the \emph{spatial} dimension, and for time embeddings to condition denoising networks over the \emph{temporal} dimension~\citep{song2020score}. Here we apply it to color channels for single pixels, in order to model fine distributional details at the level of each scalar input.

Concretely, let $z_{i,j,k}$ be the scalar value in the $k$-th channel in the $(i,j)$ spatial position of network input $\rvz_t$. We then add additional channels to the input of the denoising model of the form
\begin{align}
f^{n}_{i,j,k} = \sin(z_{i,j,k}2^{n}\pi), \text{\;\;and\;\;} g^{n}_{i,j,k} = \cos(z_{i,j,k}2^{n}\pi),
\end{align}
where $n$ runs over a range of integers $\{n_{min}, ..., n_{max}\}$. These additional channels are then concatenated to $\rvz_t$ before being used as input in a standard convolutional denoising model similar to that used by \citet{ho2020denoising}. We find that the presence of these high frequency features allows our network to learn with much higher values of $\snrmax$, or conversely lower noise levels $\sigma^{2}_{0}$, than is otherwise optimal. This leads to large improvements in likelihood as demonstrated in Section~\ref{sec:experiments} and Figure \ref{fig:fourier-ablation}. We did not observe such improvements when incorporating Fourier features into autoregressive models.

In our expreriments, we got best results with $n_{min}=7$ and $n_{max}=8$, probably since Fourier features with these frequencies are most relevant; features with lower frequencies can be learned by the network from $\rvz$, and higher frequencies are not present in the data thus irrelevant for likelihood.

\section{As a SDE}
\label{sec:sde}
When we take the number of steps $T\rightarrow\infty$, our model for $p(\rvz_t)$ can best be described as a continuous time diffusion process \citep{song2020score}, governed by the stochastic differential equation
\begin{align}
d\rvz_t = [f(t)\rvz_{t} - g^{2}(t)s_{\bT}(\rvz_t; t)] dt + g(t)d\rvW_{t},
\end{align}
with time running backwards from $t=1$ to $t=0$, where $\rvW$ denotes a Wiener process.

\section{Derivation of the VLB estimators}
\label{sec:objective}

\subsection{Discrete-time VLB}
Similar to ~\citep{sohl2015deep}, we decompose the negative variational lower bound (VLB) as:
\begin{align}
    - \log p(\rvx) \leq %
    - \text{VLB}(\rvx) =
    \underbrace{%
    D_{KL}(q(\rvz_1|\rvx)||p(\rvz_1))
    }_{\text{Prior loss}}
    + \underbrace{%
    \E_{q(\rvz_0|\rvx)}\left[ - \log p(\rvx|\rvz_0) \right]
    }_{\text{Reconstruction loss}}
    +
    \underbrace{
    \lT.
    }_{\text{Diffusion loss}}
\end{align}
The prior loss and reconstruction loss can be (stochastically and differentiably) estimated using standard techniques. We will now derive an estimator for the diffusion loss $\lT$, the remaining and more challenging term. In the case of finite $T$, using $s(i) = (i-1)/T$, $t(i) = i/T$, the diffusion loss is:
\begin{align}
    \lT
    &= 
    \sum_{i=1}^T
    \E_{q(\rvz_{t(i)}|\rvx)} 
    D_{KL}[q(\rvz_{s(i)}|\rvz_{t(i)},\rvx)||p(\rvz_{s(i)}|\rvz_{t(i)})].
\end{align}
We will use $s$ and $t$ as shorthands for $s(i)$ and $t(i)$. We will first derive an expression of $D_{KL}(q(\rvz_s|\rvz_t,\rvx)||p(\rvz_s|\rvz_t))$.

Recall that $p(\rvz_s|\rvz_t) = q(\rvz_s|\rvz_t, \rvx=\hat{\rvx}_{\bT}(\rvz_t; t))$, and thus $q(\rvz_s|\rvz_t,\rvx) = \mathcal{N}(\rvz_s; \bmu_{Q}(\rvz_{t}, \rvx; s,t), \sigma^2_Q(s,t) \bfI)$ and $p(\rvz_s|\rvz_t) = \mathcal{N}(\rvz_s; \bmu_{\bT}(\rvz_{t}; s, t), \sigma^2_Q(s,t) \bfI)$, with
\begin{align}
\bmu_{Q}(\rvz_t, \rvx; s, t) &= \frac{\alpha_{t|s}\sigma^{2}_s}{\sigma^{2}_t}\rvz_{t} + \frac{\alpha_s \sigma^{2}_{t|s}}{\sigma^{2}_{t}}\rvx\\
\bmu_{\bT}(\rvz_t; s, t) &= \frac{\alpha_{t|s}\sigma^{2}_s}{\sigma^{2}_t}\rvz_{t} + \frac{\alpha_s \sigma^{2}_{t|s}}{\sigma^{2}_{t}}\hat{\rvx}_{\bT}(\rvz_t; t),\\
\text{and\;\;} \sigma^2_Q(s,t) &= \sigma^2_{t|s}\sigma^2_s / \sigma^2_t.
\end{align}
Since $q(\rvz_s|\rvz_t,\rvx)$ and $p(\rvz_s|\rvz_t)$ are Gaussians, their KL divergence is available in closed form as a function of their means and variances, which due to their with equal variances simplifies as:
\begin{align}
D_{KL}(q(\rvz_s|\rvz_t,\rvx)||p(\rvz_s|\rvz_t)) 
&= \frac{1}{2\sigma^2_Q(s,t)}
||\bmu_Q - \bmu_\bT||_2^2
\label{eq:KL_DPM}
\\
&= 
\frac{\sigma^2_t}{2\sigma^2_{t|s}\sigma^2_s}
\frac{\alpha_s^{2} \sigma^{4}_{t|s}}{\sigma^{4}_{t}}
||\rvx - \hat{\rvx}_{\bT}(\rvz_t; t)||_2^2\\
&= 
\frac{1}{2\sigma^2_s}
\frac{\alpha_s^{2} \sigma^{2}_{t|s}}{\sigma^{2}_{t}}
||\rvx - \hat{\rvx}_{\bT}(\rvz_t; t)||_2^2\\
&= 
\frac{1}{2\sigma^2_s}
\frac{\alpha_s^{2}(\sigma^{2}_t - \alpha_{t|s}^{2}\sigma^{2}_s)}{\sigma^{2}_{t}}
||\rvx - \hat{\rvx}_{\bT}(\rvz_t; t)||_2^2\\
&= 
\frac{1}{2}
\frac{\alpha_s^{2}\sigma^{2}_t/\sigma^2_s - \alpha_{t}^{2}}{\sigma^{2}_{t}}
||\rvx - \hat{\rvx}_{\bT}(\rvz_t; t)||_2^2\\
&= 
\frac{1}{2}\left(
\frac{\alpha_s^{2}}{\sigma^2_s} - \frac{\alpha_t^{2}}{\sigma^{2}_t}\right)
||\rvx - \hat{\rvx}_{\bT}(\rvz_t; t)||_2^2\\
&= 
\frac{1}{2}\left(
\snr(s) - \snr(t)\right)
||\rvx - \hat{\rvx}_{\bT}(\rvz_t; t)||_2^2
\end{align}
Reparameterizing $\rvz_t \sim q(\rvz_t|\rvx)$ as $\rvz_t = \alpha_t \rvx + \sigma_t \bepsilon$, where $\bepsilon \sim \mathcal{N}(0,\bfI)$, our diffusion loss becomes:
\begin{align}
\mathcal{L}_T(\rvx) &= \sum_{i=1}^T \E_{q(\rvz_t|\rvx)}[ D_{KL}(q(\rvz_s|\rvz_t,\rvx)||p(\rvz_s|\rvz_t))]\\
&= \frac{1}{2}\E_{\bepsilon \sim \mathcal{N}(0,\bfI)}[\sum_{i=1}^T
\left(
\snr(s) - \snr(t)\right)
||\rvx - \hat{\rvx}_{\bT}(\rvz_t; t)||_2^2]
\end{align}

\subsection{Estimator of $\lT$}
To avoid having to compute all $T$ terms when calculating the diffusion loss, we construct an unbiased estimator of $\lT$ using
\begin{align}
\lT =
\frac{T}{2}
\E_{\bepsilon \sim \mathcal{N}(0,\bfI), i \sim U\{1, T\}}
\left[
\left(\text{SNR}(s)-\text{SNR}(t)\right)
||\rvx - \hat{\rvx}_{\bT}(\rvz_t;t) ||_2^2
\right]\end{align}
where $U\{1, T\}$ is the discrete uniform distribution from $1$ to (and including) $T$, $s = (i-1)/T$, $t = i/T$ and $\rvz_t = \alpha_t \rvx + \sigma_t \bepsilon$. This trivially yields an unbiased Monte Carlo estimator, by drawing random samples $i \sim U\{1, T\}$ and $\bepsilon \sim \mathcal{N}(0,\bfI)$.

\subsection{Infinite depth ($T \to \infty$) }\label{sec:infdepth}
To calculate the limit of the diffusion loss as $T\rightarrow\infty$, we express $\lT$ as a function of $\tau = 1/T$:
\begin{align}
\lT =
\frac{1}{2}
\E_{\bepsilon \sim \mathcal{N}(0,\bfI), i \sim U\{1, T\}}
\left[
\frac{\text{SNR}(t-\tau)-\text{SNR}(t)}{\tau}
||\rvx - \hat{\rvx}_{\bT}(\rvz_t;t) ||_2^2
\right],
\end{align}
where again $t = i/T$ and $\rvz_t = \alpha_t \rvx + \sigma_t \bepsilon$.

As $\tau \rightarrow 0, T \rightarrow \infty$, and letting $\snr'(t)$ denote the derivative of the SNR function, this then gives
\begin{align}
\linfty &=
-\frac{1}{2}
\E_{\bepsilon \sim \mathcal{N}(0,\bfI), t \sim U[0,1]}
\left[\snr'(t)||\rvx - \hat{\rvx}_{\bT}(\rvz_t;t) ||_2^2
\right]\\
&= -\frac{1}{2}\E_{\bepsilon\sim\mathcal{N}(0,\bfI)} \int_{0}^{1} \snr'(t) \left\rVert \rvx - \hat{\rvx}_{\bT}(\rvz_t;t) \right\lVert_{2}^{2} dt
.
\end{align}

\section{Influence of the number of steps $T$ on the VLB}
\label{appendix:more_steps}
Recall that the diffusion loss for our choice of model $p,q$, when using $T$ timesteps, is given by
\begin{align}
\lT =
\frac{1}{2}\E_{\bepsilon \sim \mathcal{N}(0,\bfI)}
\left[\sum_{i=1}^T
\left(\snr(s(i))-\snr(t(i))\right)
||\rvx - \hat{\rvx}_{\bT}(\rvz_{t(i)};t(i)) ||_2^2
\right],
\end{align}
with $s(i) = (i-1)/T$, $t(i) = i/T$.

This can then be written equivalently as
\begin{align}
\lT =
\frac{1}{2}\E_{\bepsilon \sim \mathcal{N}(0,\bfI)}
\left[\sum_{i=1}^T
\left(\snr(s)-\snr(t') + \snr(t')-\snr(t)\right)
||\rvx - \hat{\rvx}_{\bT}(\rvz_{t};t) ||_2^2 \right],
\end{align}
with $t' = t - 0.5/T$.

In contrast, the diffusion loss with 2T timesteps can be written as
\begin{align}
\ldT =
\frac{1}{2}\E_{\bepsilon \sim \mathcal{N}(0,\bfI)}
& \sum_{i=1}^T
\left(\snr(s)-\snr(t')\right)
||\rvx - \hat{\rvx}_{\bT}(\rvz_{t'};t') ||_2^2 \nonumber\\
&+ \sum_{i=1}^T
\left(\snr(t')-\snr(t)\right)
||\rvx - \hat{\rvx}_{\bT}(\rvz_{t};t) ||_2^2.
\end{align}
Subtracting the two results, we get
\begin{align}
&\ldT - \lT = \nonumber\\ &\frac{1}{2}\E_{\bepsilon \sim \mathcal{N}(0,\bfI)}
\left[\sum_{i=1}^T (\snr(s)-\snr(t')) \left(
||\rvx - \hat{\rvx}_{\bT}(\rvz_{t'};t') ||_2^2 - ||\rvx - \hat{\rvx}_{\bT}(\rvz_{t};t) ||_2^2 \right)\right].
\end{align}

Since $t' < t$, $\rvz_{t'}$ is a less noisy version of the data from earlier in the diffusion process compared to $\rvz_{t}$. Predicting the original data $\rvx$ from $\rvz_{t'}$ is thus strictly easier than from $\rvz_{t}$, leading to lower mean squared error if our model $\hat{\mathbf{x}}_{\bT}$ is good enough. We thus have that $\ldT - \lT < 0$, which means that doubling the number of timesteps always improves our diffusion loss. For this reason we argue for using the continuous-time VLB corresponding to $T\rightarrow\infty$ in this paper.

\section{Equivalence of diffusion specifications}
\label{sec:app_equivalence}
(Continuation of Section \ref{sec:schedule_invariance}.)

Let $p^{A}(\rvx)$ denote the distribution on observed data $\rvx$ as defined by the combination of a diffusion specification $\{\alpha^A_v, \sigma^A_v\}$ and denoising function $\tilde{\rvx}^A_{\bT}$, and let $\rvz^{A}_v$ denote the latents of this diffusion process. Similarly, let $p^{B}(\rvx), \rvz^{B}_v$ be defined through a different specification $\{\alpha^B_v, \sigma^B_v, \tilde{\rvx}^B_{\bT}\}$.

Since $v \equiv \alpha_v^2/\sigma_v^2$, we have that $\sigma_v = \alpha_v/\sqrt{v}$, which means that $\rvz_v(\rvx, \bepsilon) = \alpha_v \rvx + \sigma_v \bepsilon = \alpha_v( \rvx + \bepsilon/\sqrt{v})$. This holds for any diffusion specification by definition, and therefore we have $\rvz^A_v(\rvx, \bepsilon) = (\alpha^A_v/\alpha^B_v) \rvz_v^B(\rvx, \bepsilon)$. The latents $\rvz_v$ for different diffusion specifications are thus identical, up to a trivial rescaling, and their information content only depends on the signal-to-noise ratio $v$, not on $\alpha_v,\sigma_v$ separately.

For the purpose of denoising from a latent $\rvz^{B}_{v}$, this means we can simply define the denoising model as $\tilde{\rvx}^B_{\bT}(\rvz^{B}_{v}, v) \equiv \tilde{\rvx}^A_{\bT}((\alpha^A_v/\alpha^B_v)\rvz^{B}_{v}, v)$, and we will then get the same reconstruction loss as when denoising from $\rvz^{A}_{v}$ using model $\tilde{\rvx}^A_{\bT}$. Assuming endpoints $\snrmax$ and $\snrmin$ are equal for both specifications, \Eqref{eq:elbo_v} then tells us that $\mathcal{L}^A_{\infty}(\rvx) = \mathcal{L}^B_{\infty}(\rvx)$, i.e.\ they both produce the same diffusion loss in continuous time.

Similarly, the conditional model distributions over the latents $\rvz_{v}$ in our generative model are functions of the denoising model $\tilde{\rvx}_{\bT}(\rvz_{v}, v)$ (see Equation~\ref{eq:zs_given_zt_model}), and we therefore have that the specification $\{\alpha^A_v, \sigma^A_v, \tilde{\rvx}^A_{\bT}\}$ defines the same generative model over $\rvz_{v}, \rvx$ as the specification $\{\alpha^B_v, \sigma^B_v, \tilde{\rvx}^B_{\bT}\}$, again up to a rescaling of the latents by $\alpha^A_v/\alpha^B_v$.

\section{Implementation of monotonic neural net noise schedule $\gamma_{\eta}(t)$}
\label{app:monotonicnn}

To learn the signal-to-noise ratio $\snr(t)$, we parameterize it as $\snr(t) = \exp(-\gamma_{\boldeta}(t))$ with $\gamma_{\boldeta}(t)$ a monotonic neural network. This network consists of 3 linear layers with weights that are restricted to be positive, $l_{1},l_{2},l_{3}$, which are composed as $\tilde{\gamma}_{\boldeta}(t) = l_{1}(t)+l_{3}(\phi(l_{2}(l_{1}(t))))$, with $\phi$ the sigmoid function. Here, the $l_2$ layer has 1024 outputs, and the other layers have a single output.

In case of the continuous-time model, for the purpose of variance minimization, we postprocess the noise schedule as detailed in Appendix \ref{app:variance_minimization}.

\section{Variance minimization}
\label{app:variance_minimization}

Reduction of the variance diffusion loss estimator can lead to faster optimization. 

For the continuous-time model, we reduce the variance of the diffusion loss estimator through two methods: (1) optimizing the noise schedule w.r.t. the variance of the diffusion loss, and (2) using a low-discrepency sampler. Note that these methods can be omitted if one aims for a simple implementation of our methods, at the expense of slower optimization.

\subsection{Low-discrepency sampler}
\label{sec:lowdisc}
When processing a minibatch of $k$ examples $\rvx^{i}$, $i \in \{1,\ldots,k\}$, we require $k$ timesteps $t^{i}$ sampled from a uniform distribution. Instead of sampling these timesteps independently, we sample a single uniform random number $u_{0} \sim U[0,1]$ and then set $t^{i} = \text{mod}(u_{0} + i/k, 1)$. Each $t^{i}$ now has the correct uniform marginal distribution, but the minibatch of timesteps covers the space in $[0,1]$ more equally than when sampling independently, which we find to reduce the variance in our VLB estimate.

\subsection{Optimizing the noise schedule w.r.t. the variance of the diffusion loss}
In case of the continuous-time model, for the purpose of variance minimization, we postprocess the noise schedule as follows. At this point, the range of $\tilde{\gamma}_{\boldeta}(t)$ is unbounded and so the resulting $\snr$ is not yet restricted to $[\snrmin,\snrmax]$. We therefore postprocess the monotonic neural network as
\begin{align}
\gamma_{\boldeta}(t) = \gamma_0 + (\gamma_1 - \gamma_0)\frac{\tilde{\gamma}_{\boldeta}(t) - \tilde{\gamma}_{\boldeta}(0)}{\tilde{\gamma}_{\boldeta}(1) - \tilde{\gamma}_{\boldeta}(0)},
\end{align}
with $\gamma_0 = -\log(\snrmax), \gamma_1 = -\log(\snrmin)$. Now $\snr(t) = \exp(-\gamma_{\boldeta}(t))$ has the correct range and interpolates exactly between $\snrmin$ and $\snrmax$. We treat $\gamma_0, \gamma_1$ as free parameters that we optimize directly w.r.t. the VLB. The remaining parameters $\boldeta$ are instead learned by minimizing the variance of the stochastic estimate of the VLB.

We minimize the variance by performing stochastic gradient descent on our squared diffusion loss $\mathcal{L}^{MC}_{\infty}(\rvx,w,\gamma)^{2}$. We have that $\E_{t,\bepsilon}[\mathcal{L}^{MC}_{\infty}(\rvx,w,\gamma)^{2}] = \mathcal{L}_{\infty}(\rvx,w)^{2} + \Var_{t,\bepsilon}[\mathcal{L}^{MC}_{\infty}(\rvx,w,\gamma)]$, where the first part is independent of $\gamma_{\boldeta}(t)$, and hence that
\begin{align}
\E_{t,\bepsilon}[\nabla_{\boldeta} \mathcal{L}^{MC}_{\infty}(\rvx,w,\gamma_{\boldeta})^{2}] =    \nabla_{\boldeta} \Var_{t,\bepsilon}[\mathcal{L}^{MC}_{\infty}(\rvx,w,\gamma_{\boldeta})].
\end{align}
We can calculate this gradient with negligible computational overhead as a by-product of calculating the gradient of the VLB, details of which are given in Appendix~\ref{app:monotonicnn}.

We wish to calculate $\nabla_{\boldeta}[ \mathcal{L}^{MC}_{\infty}(\rvx,\gamma_{\boldeta})^{2}]$ without performing a second backpropagation pass through the denoising model due to this objective being different than for the other parameters. To do this, we decompose the gradient as
\begin{align}
\frac{d}{d \boldeta}[ \mathcal{L}^{MC}_{\infty}(\rvx,\gamma_{\boldeta})^{2}] &= \frac{d}{d \snr}\left[ \mathcal{L}^{MC}_{\infty}(\rvx,\snr)^{2}\right]\frac{d}{d \boldeta}\left[\snr(\boldeta)\right],\\
\text{and\;\;} \frac{d}{d \snr}\left[ \mathcal{L}^{MC}_{\infty}(\rvx,\snr)^{2}\right] &= 2\frac{d}{d \snr}\left[ \mathcal{L}^{MC}_{\infty}(\rvx,\snr)\right] \odot \mathcal{L}^{MC}_{\infty}(\rvx,\snr),
\end{align}
where $\odot$ denotes elementwise multiplication.
Here $\frac{d}{d \snr}\left[ \mathcal{L}^{MC}_{\infty}(\rvx,\snr)\right]$ is computed along with the other gradients when performing the single backpropagation pass for calculating $\nabla_{\bT}[ \mathcal{L}^{MC}_{\infty}]$. The remaining operations required to get $\nabla_{\boldeta} [\mathcal{L}^{MC}_{\infty}(\rvx,\gamma_{\boldeta})^{2}]$ have negligible computational cost.

This strategy of minimizing the variance of our diffusion loss estimate remains valid for weighted diffusion losses, $w(v) \neq 1$, not corresponding to the VLB, and we therefore expect it to be useful beyond the goal of optimizing for likelihood that we consider in this paper.

\section{Numerical stability}
\label{sec:stable}

Floating point numbers are much worse at representing numbers close to 1, than at representing numbers close to 0. Since a na\"ive implementation of our model and its discrete-time loss function requires computing intermediate values that are close to 1, those numbers are erroneously rounded to 1, leading to numerical issues and incorrect results. Note that previous implementations of discrete-time diffusion models (e.g. ~\citep{ho2020denoising}) used 64-bit floating point numbers to avoid numerical problems. We found this unnecessary in our model.

A numerically problematic term, for example, is $\bsst^2$ which is used for sampling. It is straightforward to verify that:
\begin{align}
\bsst^2 &= -\text{expm1}(\text{softplus}(\gamma(s))-\text{softplus}(\gamma(t))),
\end{align}
where $\text{expm1}(x) \equiv \exp(x) - 1$ and $\text{softplus}(x) \equiv \log(1 + \exp(x))$ are functions with numerically stable primitives in common numerical computing packages. 

\section{Comparison to DDPM and NCSN objectives}
\label{sec:ncsn_ddpm_weights}
Previous works using denoising diffusion models \citep{ho2020denoising, song2019generative, nichol2021improved} used a training objective that can be understood as a \emph{weighted} diffusion loss of the form given in \Eqref{eq:weighted_diff}:
\begin{align}
\mathcal{L}_{\infty}(\rvx, w) &= \frac{1}{2}\E_{\bepsilon\sim\mathcal{N}(0,\bfI)} \int_{\snr_{\text{min}}}^{\snr_{\text{max}}} w(v) \left\rVert \rvx - \tilde{\rvx}_{\bT}(\rvz_v, v) \right\lVert_{2}^{2} dv \label{eq:weighted_diff_app}\\
&= -\frac{1}{2}\E_{\bepsilon\sim\mathcal{N}(0,\bfI)} \int_{0}^{1} \snr'(t)w(\snr(t)) \left\rVert \rvx - \hat{\rvx}_{\bT}(\rvz_t;t) \right\lVert_{2}^{2} dt\\
&= \frac{1}{2}\E_{\bepsilon\sim\mathcal{N}(0,\bfI)} \int_{0}^{1} \gamma'(t)w(\exp(-\gamma(t))) \left\rVert \bepsilon - \hat{\bepsilon}_{\bT}(\rvz_t;t) \right\lVert_{2}^{2} dt,\label{eq:weighted_gamma_loss}
\end{align}
where $\gamma(t)=-\log\snr(t)$.

When using the loss in \Eqref{eq:weighted_diff_app}, we set $w(v)=1$, corresponding to optimization of a variational bound on the likelihood of the data. \cite{ho2020denoising, song2019generative,nichol2021improved} instead choose to minimize the \emph{simple objective} defined as
\begin{align}
L_{\text{simple}}(\rvx) \equiv \int_{0}^{1} \lVert \bepsilon - \hat{\bepsilon}_{\bT}(\rvz_{t}, t)\rVert_{2}^{2} dt,
\label{eq:simple_loss}
\end{align}
or a discrete-time version of this.

\begin{figure}[htb]
    \centering
    \includegraphics[width=0.5\textwidth]{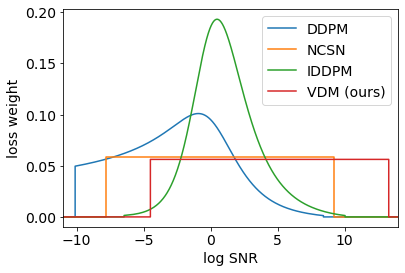}
    \caption{Implied weighting functions corresponding to the losses used by \cite{ho2020denoising}, \cite{song2020denoising}, and \cite{nichol2021improved}, as well as our proposed loss. NCSN \citep{song2020denoising} uses a constant implied weighting function, and is thus consistent with maximization of the variational bound like we propose in this paper. However, unlike \cite{song2019generative} we also learn the endpoints $\snrmin,\snrmax$, which results in a better optimized VLB value. DDPM~\citep{ho2020denoising} and improved DDPM~\citep{nichol2021improved} instead use implied weighting functions that put relatively more weight on the noisy data with low to medium signal-to-noise ratio. The latter two works report better FID and Inception Score than \cite{song2020denoising} and the current paper, which we hypothesize is due to their loss emphasizing the global consistence and coarse level patterns more than the fine scale features of the data.}
    \label{fig:weighting_funs}
\end{figure}

Comparing \Eqref{eq:simple_loss}
with \Eqref{eq:weighted_gamma_loss}, we can see that the loss used by \cite{ho2020denoising,song2019generative,nichol2021improved} corresponds to a weighting function $w(\snr(t))=1/\gamma'(t)$. Below, we derive the $\gamma(t)$, and thus the weighting function $w(v)$, corresponding to the diffusion processes used by \cite{ho2020denoising,song2019generative,nichol2021improved}. We visualize these weighting functions in Figure~\ref{fig:weighting_funs}.

\begin{figure}[htb]
    \centering
    \includegraphics[width=0.5\textwidth]{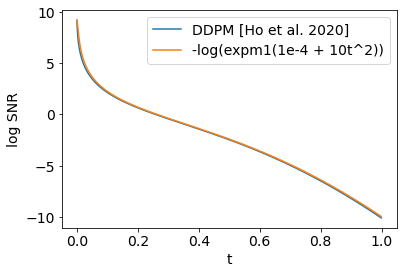}
    \caption{Log signal-to-noise ratio for the discrete-time diffusion process in \cite{ho2020denoising} and our continous-time approximation.}
    \label{fig:ho_snr}
\end{figure}

For DDPM, \cite{ho2020denoising} use a diffusion process in discrete time with $\alpha_i = \sqrt{\prod_{j=1}^{i}(1-\beta_j)}$, $\sigma^{2}_i = 1 - \alpha^{2}_i$, where $\beta_i$ linearly interpolates between $\beta_{1}=1e^{-4}$ and $\beta_{T}=0.02$ in $T=1000$ discrete steps. When defining time $t = i/T$, this can be closely approximated as $\alpha^{2}_t = \exp(-1e^{-4} - 10t^{2})$, and correspondingly with $\snr(t) = 1/\text{expm1}(1e^{-4} + 10t^{2})$ or $\gamma(t) = \log[\text{expm1}(1e^{-4} + 10t^{2})]$, where $\text{expm1}(x) = \exp(x)-1$. This approximation is shown in Figure~\ref{fig:ho_snr}.

For NCSNv2, \cite{song2020improved} instead use $\alpha_t = 1$ and let $\sigma_{t}$ be a geometric series interpolating between $0.01$ and $50$, i.e. $\sigma^{2}_t = \exp(\gamma(t))$ with $\gamma(t)=2\log[0.01] + 2\log[5000]t$. This means that $\gamma'(t)=2\log[5000]$ and thus that $w(v)$ is a constant. The procedure proposed by \cite{song2020improved} is thus consistent with maximization of the VLB like we propose here. The same holds for ~\citep{song2019generative}.

For IDDPM, \cite{nichol2021improved} use $\tilde{\alpha}_{t} = \cos(\frac{t+0.008}{1.008}\frac{\pi}{2})/\cos(\frac{0.008}{1.008}\frac{\pi}{2})$. The values for $\tilde{\alpha}_{t}$ are then translated into value for $\beta_t$, which are then clipped to $0.999$. Subsequently we can then derive the $\alpha_{t},\sigma_t,\gamma(t)$ corresponding to those $\beta_t$. Due to the clipping these expressions do not simplify, but we include their numerical results in Figure~\ref{fig:weighting_funs}.

\section{Consistency}
\label{sec:dsm}

Let $q(\rvx)$ denote the marginal distribution of data $\rvx$, and let:
\begin{align}
    q(\rvz_t) = \int q(\rvz_t|\rvx) q(\rvx) d\rvx
\end{align}
Here we will show that derived estimators are \emph{consistent} estimators, in the sense that with infinite data, the optimal score model $\snT^*(\rvz_t;t)$ is such that:
\begin{align}
    \snT^*(\rvz_t;t) = \nabla_{\rvz} \log q(\rvz_t)
\end{align}

Note that $\nabla_{\rvz_t} \log q(\rvz_t|\rvx) = -\bepsilon/\bsigma_t$. We can rewrite the diffusion loss (discrete time or continuous time) for timestep $t$ as:
\begin{align}
\mathcal{L}_T(\rvx; t) &=
\frac{1}{2} 
\E_{q(\rvx,\rvz_t)} \left[
\norm{
\sqrt{\rvc(t)}
    \left(\nabla_{\rvz_t} \log q(\rvz_t|\rvx) - \snT(\rvz_t; t)\right)
}_2^2
\right]
\end{align}
where $\sqrt{\rvc(t)}$ is a time-dependent weighting factor. 

In ~\citep{ho2020denoising}, it is noted that the discrete-time VLB, when using equal variances across dimensions, is equivalent to a Denoising Score Matching (DSM) objective \citep{vincent2011connection}. This is interesting, since it implies consistency. We generalize this original consistency proof of DSM to a more general case of different noises schedules per dimension, and arbitrary multipliers $\sqrt{\rvc_1}$ and $\sqrt{\rvc_2}$ in front of the scores, i.e. where the dimensions of $\rvz$ are differently weighted. Note, however that we'll only need the special case where $\sqrt{\rvc}=\sqrt{\rvc_1}=\sqrt{\rvc_2}$. 
First, note that:
\begin{align}
\tfrac{1}{2} \E_{q(\rvz_t)} \left[ ||\sqrt{\rvc_1}\nabla_{\rvz_t} \log q(\rvz_t) - \sqrt{\rvc_2}\snT(\rvz_t)||_2^2 \right]
&= \tfrac{1}{2} \E_{q(\rvz_t)} \left[ ||\sqrt{\rvc_1}\nabla_{\rvz_t} \log q(\rvz_t)||_2^2 \right]
\nonumber\\
&+ \tfrac{1}{2} \E_{q(\rvz_t)} \left[ ||\sqrt{\rvc_2}\snT(\rvz_t)||_2^2 \right]
\nonumber\\
&- \E_{q(\rvz_t)} \left[ \langle \sqrt{\rvc_1} \nabla_{\rvz_t} \log q(\rvz_t), \sqrt{\rvc_2}\snT(\rvz_t) \rangle \right]
\label{eq:dsm_a}
\end{align}
Where $\langle ., . \rangle$ denotes a dot product. Similarly:
\begin{align}
\tfrac{1}{2} \E_{q(\rvz_t|\rvx)} \left[ ||\sqrt{\rvc_1}\nabla_{\rvz_t} \log q(\rvz_t|\rvx) - \sqrt{\rvc_2}\snT(\rvz_t)||_2^2 \right]
&= \tfrac{1}{2} \E_{q(\rvz_t|\rvx)} \left[ ||\sqrt{\rvc_1}\nabla_{\rvz_t} \log q(\rvz_t|\rvx)||_2^2 \right]
\nonumber\\
&+ \tfrac{1}{2} \E_{q(\rvz_t|\rvx)} \left[ ||\sqrt{\rvc_2}\snT(\rvz_t)||_2^2 \right]
\nonumber\\
&- \E_{q(\rvz_t|\rvx)} \left[ \langle \sqrt{\rvc_1}\nabla_{\rvz_t} \log q(\rvz_t|\rvx), \sqrt{\rvc_2}\snT(\rvz_t) \rangle \right]
\label{eq:dsm_b}
\end{align}
The second terms of the right-hand sides of \Eqref{eq:dsm_a} and \Eqref{eq:dsm_b} are equal. The third terms of the right-hand sides of \Eqref{eq:dsm_a} and \Eqref{eq:dsm_b} are also equal:
\begin{align}
&\E_{q(\rvz_t)} \left[ \langle \sqrt{\rvc_2}\snT(\rvz_t), \sqrt{\rvc_1}\nabla_{\rvz_t} \log q(\rvz_t) \rangle \right]\\
&= 
\E_{q(\rvz_t)} \left[ \left\langle \sqrt{\rvc_2} \snT(\rvz_t), \sqrt{\rvc_1} \frac{\nabla_{\rvz_t} q(\rvz_t)}{q(\rvz_t)} \right\rangle \right]\\
&= \int_{\rvz_t} \left\langle \sqrt{\rvc_2}\snT(\rvz_t), \sqrt{\rvc_1}\nabla_{\rvz_t} q(\rvz_t) \right\rangle d\rvz_t\\
&= \int_{\rvz_t} \left\langle \sqrt{\rvc_2}\snT(\rvz_t), \sqrt{\rvc_1}\nabla_{\rvz_t} \int_{\rvx} q(\rvx) q(\rvz_t|\rvx) d\rvx \right\rangle d\rvz_t\\
&= \int_{\rvz_t} \left\langle \sqrt{\rvc_2}\snT(\rvz_t), \sqrt{\rvc_1}\int_{\rvx} q(\rvx) \nabla_{\rvz_t} q(\rvz_t|\rvx) d\rvx \right\rangle d\rvz_t\\
&= \int_{\rvz_t} \left\langle \sqrt{\rvc_2}\snT(\rvz_t), \sqrt{\rvc_1}\int_{\rvx} q(\rvx) q(\rvz_t|\rvx) \nabla_{\rvz_t} \log q(\rvz_t|\rvx) d\rvx \right\rangle d\rvz_t\\
&= \int_{\rvx} \int_{\rvz_t} q(\rvx) q(\rvz_t|\rvx) \left\langle \sqrt{\rvc_2}\snT(\rvz_t), \sqrt{\rvc_1} \nabla_{\rvz_t} \log q(\rvz_t|\rvx) \right\rangle d\rvz_t d\rvx\\
&= \E_{q(\rvx,\rvz_t)} \left[ \left\langle \sqrt{\rvc_2}\snT(\rvz_t),  \sqrt{\rvc_1}\nabla_{\rvz_t} \log q(\rvz_t|\rvx) \right\rangle \right]
\label{eq:dsm_c}
\end{align}
So, only the first term of the right-hand sides of \Eqref{eq:dsm_a} and \Eqref{eq:dsm_b} are not equal. It follows that:
\begin{align}
\tfrac{1}{2} \E_{q(\rvz_t)} \left[ ||\sqrt{\rvc_1}\nabla_{\rvz_t} \log q(\rvz_t) - \sqrt{\rvc_2}\snT(\rvz_t)||_2^2 \right] &= \tfrac{1}{2} \E_{q(\rvx,\rvz_t)} \left[ ||\sqrt{\rvc_1}\nabla_{\rvz_t} \log q(\rvz_t|\rvx) - \sqrt{\rvc_2}\snT(\rvz_t)||_2^2 \right]\nonumber\\
&+ \text{constant}
\label{eq:dsm}\end{align}
where $\text{constant} = \tfrac{1}{2} \E_{q(\rvz_t)} \left[ ||\sqrt{\rvc_1}\nabla_{\rvz_t} \log q(\rvz_t)||_2^2 \right] - \tfrac{1}{2} \E_{q(\rvx,\rvz_t)} \left[ ||\sqrt{\rvc_1}\nabla_{\rvz_t} \log q(\rvz_t|\rvx)||_2^2 \right]$ is constant w.r.t. the energy-based model (EBM) $E()$. In the special case where $\sqrt{\rvc_1}=\sqrt{\rvc_2}$, we have:
\begin{align}
\tfrac{1}{2} \E_{q(\rvz_t)} \left[ ||\sqrt{\rvc_1}(\nabla_{\rvz_t} \log q(\rvz_t) - \snT(\rvz_t))||_2^2 \right] &= \tfrac{1}{2} \E_{q(\rvx,\rvz_t)} \left[ ||\sqrt{\rvc_1}(\nabla_{\rvz_t} \log q(\rvz_t|\rvx) - \snT(\rvz_t))||_2^2 \right]\nonumber\\
&+ \text{constant}
\label{eq:dsm2}\end{align}
Therefore, minimizing the first term on the right-hand side of \Eqref{eq:dsm2} w.r.t. $E()$ (a denoising score matching objective with differently weighted dimensions) is equivalent to minimizing the left-hand side of \Eqref{eq:dsm2} w.r.t. $E()$. 
From this equation, it is clear that at the optimum of this DSM objective, for any positive $\rvc_1$:
\begin{align}
    \snT^*(\rvz_t) = \nabla_{\rvz_t} \log q(\rvz_t)
\end{align}
If the score model is parameterized as the gradient of an EBM $E()$, then this implies that for all $t \in [0,1]$:
\begin{align}
    \exp(-E^*(\rvz_t; t)) \propto q(\rvz_t)
\label{eq:consistency}\end{align}
So, when optimizing for the diffusion loss, the EBM $E(.;t)$ will approximate the correct marginals corresponding the inference model.

\section{Additional samples from our models}
\label{sec:app_more_samples}
\begin{figure}[htb]
    \centering
    \includegraphics{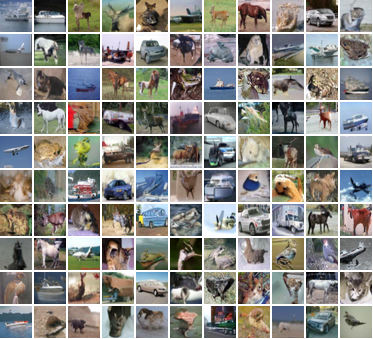}
    \caption{Random samples from an unconditional diffusion model trained on CIFAR-10 for 2 million parameter updates. The model was trained in continuous-time, and sampled using $T=1000$.}
    \label{fig:extracifar}
\end{figure}

\begin{figure}[htb]
    \centering
    \includegraphics{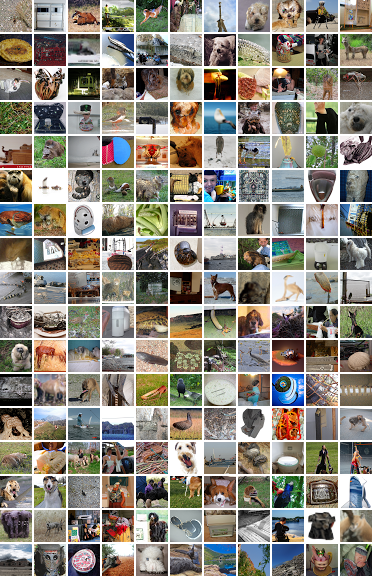}
    \caption{Random samples from an unconditional diffusion model trained on 32x32 ImageNet for 3.7 million parameter updates. The model was trained in continuous-time, and sampled using $T=1000$.}
    \label{fig:in32}
\end{figure}

\begin{figure}[htb]
    \centering
    \includegraphics{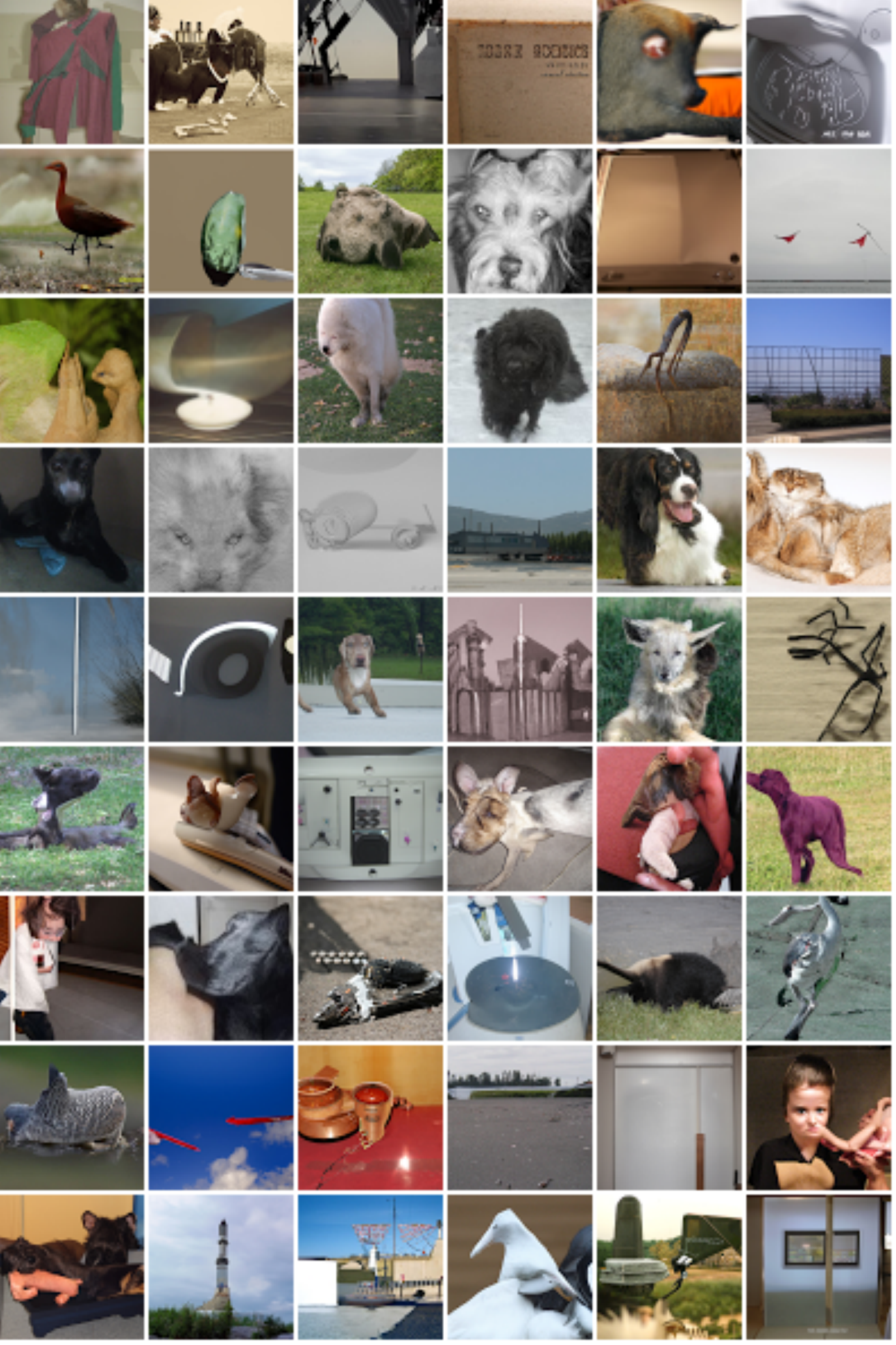}
    \caption{Random samples from an unconditional diffusion model trained on 64x64 ImageNet for 2 million parameter updates. The model was trained in continuous-time, and sampled using $T=1000$.}
    \label{fig:in64}
\end{figure}

We include additional uncurated random samples from our unconditional models trained on CIFAR-10, 32x32 Imagenet, and 64x64 Imagenet. See Figures~\ref{fig:extracifar}, \ref{fig:in32}, and \ref{fig:in64}.

\section{Lossless compression}
\label{sec:compression}
For a fixed number of evaluation timesteps $T_{eval}$, our diffusion model in discrete time is a hierarchical latent variable model that can be turned into a lossless compression algorithm using bits-back coding~\citep{hinton1993keeping}. Assuming a source of auxiliary random bits is available alongside the data, bits-back coding encodes a latent and data together, with the latent sampled from the approximate posterior using the auxiliary random bits. The net coding cost of bits-back coding is given by subtracting the number of bits needed to sample the latent from the number of bits needed to encode the latent and data using the reverse process, so the negative VLB of our discrete time model is the theoretical expected coding cost for bits-back coding.

As a proof of concept for lossless compression using our model, Table~\ref{table:t_results} reports net codelengths on the CIFAR10 test set for various settings of $T_{eval}$ using BB-ANS~\citep{townsend2018practical}, a practical implementation of bits-back coding based on asymmetric numeral systems~\citep{duda2009asymmetric}. Since diffusion models have Markov forward and reverse processes, we use the Bit-Swap implementation of BB-ANS~\citep{kingma2019bit}. Practical implementations of bits-back coding must discretize continuous latent variables and their associated continuous probability distributions; for simplicity, our implementation uses a uniform discretization of the continuous latents and their associated Gaussian conditionals from the forward and reverse processes. Additionally, we found it crucial to encrypt the ANS bitstream before each decoding operation to ensure clean bits for sampling from the approximate posterior; we did so by applying the XOR operation to the ANS bitstream with pseudorandom bits from a fixed sequence of seeds. For example, without cleaning the bitstream using encryption, compressing a batch of 100 examples using $T_{eval}=250$ costs 2.74 bits per byte, but with encryption, the cost improves to 2.68 bits per dimension.

For a small number of timesteps $T_{eval}$, our bits-back implementation attains net codelengths that agree closely with the negative VLB, but there is some discrepancy for large $T_{eval}$. This is due to inaccuracies in the compression algorithm to represent discretized Gaussians with small standard deviations, and small discrepancies in codelength  compound into a gap of up to 0.05 bits per dimension when $T$ is large. (In prior work, e.g. \citep{kingma2019bit,ho2019compression,townsend2020hilloc}, practical implementations of bits-back coding have been tested on latent variable models with only tens of layers, not hundreds.) In addition, a large number of timesteps makes compression computationally expensive, because a neural network forward pass must be run for each timestep. Closing the codelength gap with an efficient implementation of bits-back coding for a large number of timesteps is an interesting avenue for future work.

\section{Density estimation on additional data sets}
At the request of one of the reviewers we also ran our model on additional data sets of higher resolution, less diverse, images. Specifically, we obtain a test set likelihood of $2.14$ bits per dim on CelebA-HQ \citep{karras2017progressive}, and $1.44$ on LSUN bedrooms \citep{yu2015lsun}, both at $128 \times 128$ resolution. Since these are not established benchmarks in density estimation, and since downsampling methods in the literature are not consistent, we don't compare against previous methods for these data sets. Our results are provided purely to give a ballpark estimate of how well our proposed method scales to higher resolution images.

The model used for these data sets is based on that used for Imagenet $64 \times 64$, with an additional level in the UNet at resolution $128 \times 128$, consisting of 16 residual layers using 128 channels. Our model downsamples between the $128 \times 128$ and $64 \times 64$ resolutions, similar to e.g.\ \cite{ho2020denoising}, but unlike the models we used for the other data sets we considered.

\end{document}

%% file: 2_defs_general.tex
\def\eqref#1{equation~\ref{#1}}
\def\Eqref#1{Equation~\ref{#1}}

\def\1{\bm{1}}

\def\rvc{{\mathbf{c}}}

\def\rvx{{\mathbf{x}}}
\def\rvy{{\mathbf{y}}}
\def\rvz{{\mathbf{z}}}

\DeclareMathAlphabet{\mathsfit}{\encodingdefault}{\sfdefault}{m}{sl}
\SetMathAlphabet{\mathsfit}{bold}{\encodingdefault}{\sfdefault}{bx}{n}

\newcommand{\E}{\mathbb{E}}

\newcommand{\sigmoid}{\sigma}

\newcommand{\Var}{\mathrm{Var}}

%% file: 3_defs_custom.tex
\usepackage[framemethod=TikZ]{mdframed}

\newcommand{\bT}{{\boldsymbol{\theta}}}

\newcommand{\boldeta}{{\boldsymbol{\eta}}}

\newcommand{\snT}{\mathbf{s}_{\bT}}

\newcommand{\norm}[1]{\left\lVert#1\right\rVert}

\newcommand{\bfI}{\mathbf{I}}

\newcommand{\bepsilon}{{\boldsymbol{\epsilon}}}
\newcommand{\bsigma}{{\boldsymbol{\sigma}}}
\newcommand{\bmu}{{\boldsymbol{\mu}}}

\makeatletter
\usepackage{xspace} 
\def\@onedot{\ifx\@let@token.\else.\null\fi\xspace}
\DeclareRobustCommand\onedot{\futurelet\@let@token\@onedot}

\usepackage{amsmath}